\documentclass{article}

\PassOptionsToPackage{numbers, square}{natbib}

\usepackage[main,final]{neurips_2026}

\usepackage[utf8]{inputenc} %
\usepackage[T1]{fontenc}    %

\usepackage{booktabs}       %
\usepackage{amsfonts}       %
\usepackage{nicefrac}       %
\usepackage{microtype}      %
\usepackage{xcolor}         %
\usepackage{graphicx}
\usepackage{amsmath, amssymb, amsthm}
\usepackage{pgfplots}
\pgfplotsset{compat=1.18}
\usepackage{tcolorbox}
\usepackage{float}
\usepackage{booktabs}
\usepackage{multirow}
\usepackage{xspace}
\usepackage{pgfplots}
\usepackage{placeins}
\usepackage{wrapfig}
\usepgfplotslibrary{groupplots}
\usetikzlibrary{calc}
\pgfplotsset{compat=1.18}

\usepackage{hyperref}       %
\usepackage{url}            %
\hypersetup{hypertexnames=false}
\usepackage[capitalize,nameinlink]{cleveref}
\usepackage{wrapfig}
\usepackage{algorithm}
\usepackage{algpseudocode}

\Crefname{section}{Sec.}{Secs.}
\Crefname{definition}{Def.}{Defs.}
\Crefname{proposition}{Prop.}{Props.}
\Crefname{lemma}{Lem.}{Lems.}
\Crefname{assumption}{Assump.}{Assumps.}
\Crefname{figure}{Fig.}{Figs.}
\Crefname{equation}{Eq.}{Eqs.}
\Crefname{algorithm}{Alg.}{Algs.}
\Crefname{table}{Tab.}{Tabs.}
\Crefname{table}{Tab.}{Tabs.}
\Crefname{appendix}{App.}{Apps.}

\newcommand{\EDITORS}{EDITORS\xspace}
\newcommand{\authors}{a}
\newcommand{\contentd}{C_a}
\newcommand{\styled}{S_a}
\newcommand{\parad}{T_a}
\newcommand{\textspace}{\mathcal{T}}

\usepackage{siunitx}
\let\ms\relax

\newcommand{\squishlist}{
 \begin{list}{\labelitemi}
  { \setlength{\itemsep}{5pt}
     \setlength{\parsep}{0pt}
     \setlength{\topsep}{0pt}
     \setlength{\partopsep}{0pt}
     \setlength{\leftmargin}{0.7em}
     \setlength{\labelwidth}{0.5em}
     \setlength{\labelsep}{0.2em} } }

\newcommand{\squishlisttwo}{
 \begin{list}{\labelitemi}
  { \setlength{\itemsep}{2pt}
     \setlength{\parsep}{1pt}
    \setlength{\topsep}{2pt}
    \setlength{\partopsep}{0pt}
    \setlength{\leftmargin}{1em}
    \setlength{\labelwidth}{1.5em}
    \setlength{\labelsep}{0.5em} } }

\newcommand{\squishend}{
  \end{list}  }

\title{\EDITORS Know Your Style! \underline{Edit}ing L\underline{oR}A \underline{S}ubspaces for Stylistic Attribution and Imitation}

\author{%
  Zixuan Wang\thanks{These authors contributed equally.}$\ ~^{1\,2}$
  \quad
  Gregory Kang Ruey Lau\footnotemark[1]$\ ~^{1\,2}$
  \quad
  Xiao Tian$^{1\,4}$
  \quad
  \textbf{Jue Fan}$^{1\,4}$ \\
  \textbf{Caroline Chaux}$^3$
  \quad
  \textbf{Bryan Kian Hsiang Low}$^1$ \\
  $^1$School of Computing, National University of Singapore, Singapore \\
  $^2$CNRS@CREATE, 1 Create Way, \#08-01 Create Tower, Singapore 138602 \\
  $^3$Aix Marseille Univ, CNRS, I2M, Marseille, France \\
  $^4$Agency for Science, Technology and Research (A*STAR), Singapore \\
  \texttt{\{e0492448, gregorylau, xiao.tian, jue.fan\}@u.nus.edu} \\
  \texttt{caroline.chaux@cnrs.fr}, \texttt{dcslkh@nus.edu.sg}
}

\begin{document}

\maketitle

\begin{abstract}
Disentangling writing style from semantic content is a fundamental challenge in literary text modeling. Style-content entanglement causes models to rely on semantic content for \textit{authorship attribution} (AA) tasks and memorize author-specific content for \textit{style imitation} (SI) tasks. We propose \EDITORS (\textbf{Edit}ing L\textbf{oR}A \textbf{S}ubspaces), a \textit{diagnose-then-deflate} framework that adapts pretrained large language models and prioritizes stylistic features, making it effective for both AA and SI tasks. \EDITORS first trains a diagnostic LoRA adapter using style-neutral statements derived from the training corpus to identify a content-oriented subspace. It then trains the final style adapter under activation-space deflation that projects input activations away from the identified content-oriented directions. We empirically demonstrate that \EDITORS is effective. On three AA benchmarks spanning literature, social media, and news, \EDITORS achieves state-of-the-art accuracy, outperforming the strongest existing baseline by up to 29\%. On SI, \EDITORS generates stylistically faithful and diverse passages, outperforming baselines in achieving style similarity while adhering to content specifications. Our code is available at \url{https://github.com/zixuan-n/editors}.
\end{abstract}

\section{Introduction} \label{sec:introduction}
The disentanglement of style and content has been extensively studied in images \citep{huang2018multimodalunsupervisedimagetoimagetranslation,kwon2021clipstyler,mou2023t2i,preechakul2021diffusion,sohn2023styledrop}, yet its generalization to literary text remains a complex challenge: style and content are deeply entangled in natural languages. The disentanglement is crucial in the era of large language models (LLMs) due to two important real-world applications:  \textit{authorship attribution} (AA), which identifies the most likely author of a given text, and  \textit{style imitation} (SI), which generates text in the style of a target author and preserves user-provided content.
However, in AA, current methods either (a) conflate style with content \citep{fabien-etal-2020-bertaa, huertastato2023understandingwritingstylesocial, huertastato2025partpretrainedauthorshiprepresentation, wang2024tracetransformerbasedattributionusing} %
due to their over-reliance on content and fail when an author writes across topics or when attribution must generalize across domains, or (b) aggressively remove useful content information \citep{chen2021disentangledcontrastivelearninglearning,contrastiveAA, huertastato2024isolatingauthorshipcontentsemantic, patel-etal-2025-styledistance, wegmann-etal-2022-author}. Likewise, in SI, existing methods (a) risk memorizing training content and provide limited support for content customization \citep{lau-etal-2018-deep, liu-etal-2024-customizing-large, kar94992}, or (b) forcefully avoid content %
with expensive data pairs and only perform narrow paragraph-to-paragraph rewriting \citep{grondahl2020effectivewritingstyleimitation, liu2024authorshipstyletransferpolicy,syed2020stylelm,zhu-etal-2023-storytrans}. 
These approaches reveal a persistent trade-off: retaining content can improve task performance but risks shortcut learning and memorization, while ignoring content can discard information that remains useful for attribution and generation.

To tackle the challenge of balancing style and content in both AA and SI, this paper presents \textbf{\EDITORS} (\textbf{Edit}ing L\textbf{oR}A \textbf{S}ubspaces), a framework that uses LoRA adapters to diagnose and suppress content-oriented directions during supervised fine-tuning (SFT)\footnote{\EDITORS can be applied to any SFT method besides LoRA (\cref{app:lora}), but we focus on LoRA in this paper because it is efficient and simplifies our computation. See \cref{app:efficient_svd_for_lora} for more explanations and demonstration on non-LoRA methods.} to \textbf{encourage stylistic features while retaining useful content information}. Our framework is motivated by the intuition that style and content are partially, but not fully, separated in the model's activation space. By diagnosing the content-oriented subspace and projecting away from it during style learning, we can train a style adapter which is encouraged to rely on stylistic features (\cref{sec:geometric-interventions}). Based on this, \EDITORS adopts a \textit{diagnose-then-deflate} strategy: 
We first \textbf{diagnose important content-oriented directions} using a set of \textit{style-neutral statements} (\cref{sec:method_diagnosis}).
We then \textbf{prioritize learning stylistic features} by suppressing these selected content-oriented directions through \textbf{activation-space \textit{deflation}} (i.e., projecting training data away from them) (\cref{sec:method_deflation}). 
Our framework \EDITORS can be applied to both AA and SI tasks with slight adjustments (\cref{sec:method_task_training}).
Our specific contributions include:
\squishlisttwo
\item We propose \EDITORS, a diagnose-then-deflate framework that prioritizes style learning by suppressing selected content directions in activation space (\cref{sec:method});
\item On AA, \EDITORS consistently improves over baselines across various domains (\cref{sec:aa_experiments});
\item On SI, \EDITORS achieves \textbf{higher style similarity, lower memorization of training content and better preservation of user-provided content}, demonstrating that the learned author-specific stylistic features transfer beyond direct copying (\cref{sec:style_imitation}).
\squishend

\vspace{-1mm}

\section{Problem Formulation}
\label{sec:problem_formulation}

Let \(\mathcal{A}\) be a finite set of authors, \(\mathcal{C}\) be the
space of semantic content specifications, and $\textspace$ be the space of text passages. We consider the setting where each author
\(a\in\mathcal{A}\) is associated with two distributions: a \textbf{content}
distribution \(\contentd\) over \(\mathcal{C}\) that captures what author $a$ tends
to write about, and an expression distribution \(\styled(\cdot\mid c)\)
over text passages $t \in \textspace$ that captures how the author expresses a given content
specification \(c \in \mathcal{C}\) as text, which we associate with author \textbf{style}. 
The text passage distribution over $\textspace$ that we can observe for author $\authors$ is then the marginal
$ \parad(t)=\int_{\mathcal{C}} \styled(t\mid c)\,d\contentd(c)$.
Text samples $t \in \textspace$ from \(\parad\) entangle two author-specific signals: the author's
typical content \(\contentd\) and style of realizing content \(\styled(\cdot\mid c)\). In this work, given a training dataset $\mathcal{D} =\{(t_i,a_i)\}_{i=1}^{N}$ covering all authors $a \in \mathcal{A}$, our goal is to develop a framework to address such entanglement and prioritize stylistic features when fine-tuning LLMs, which is essential to tasks including AA and SI.

\textbf{Authorship attribution (AA).}
In the closed-set AA task, given a passage $t$ and a
candidate author set \(\mathcal{A}\), our goal is to predict the most likely author
$ \hat{a} = \arg\max_{a\in\mathcal{A}} P(a\mid t)$. We aim to do so by training an attribution model based on training data $\mathcal{D} =\{(t_i,a_i)\}_{i=1}^{N}$, so that it can perform well on held-out passages $\mathcal{D}_\text{test}$ too.
Ideally, the %
model should identify which author-specific marginal distribution \(\parad\) most likely produced the passage, while avoiding brittle reliance on content shortcuts from \(\contentd\). This distinction matters because different authors may write about similar content, and the same author may write about different content across topics,
genres, or domains unseen in training.

\textbf{Style imitation (SI).}
In the general SI task, the model is given a target author $a \in \mathcal{A}$ and the goal is to generate a passage $\hat{t} \sim P(\cdot | a)$ that approximates author $a$'s marginal distribution $T_a$. A useful variant is \textit{content-controlled SI}, 
where the model is additionally given a content specification \(\ddot{c}\in\mathcal{C}\). The goal is to generate a passage
$\hat{t}
    \sim
    P(\cdot\mid a,\ddot{c})$
that approximates author $a$'s expression distribution \(\styled(\cdot\mid \ddot{c})\), or essentially how the author would likely express the requested content $\ddot{c}$. The ideal generative model should (1) be stylistically faithful to the target author $a$, (2) accurately convey the specified content $\ddot{c}$, and (3) avoid just copying from its training data $\mathcal{D}$.

Note that both tasks require learning from samples $t \sim \parad$, where content $\contentd$ and style $\styled$ are observed entangled in text passage form. For AA, content information from \(\contentd\) can be
useful but misleading: some content patterns may help identify an
author easily during training, but over-reliance on \(\contentd\) can lead to poor
generalization under topic or domain shift (e.g., when an author writes about a different topic). For SI, the issue may
be more problematic: preserving author-specific content patterns from \(\contentd\) can cause
the model to reproduce training
passages or ignore the user-provided content specification \(\ddot{c}\) in content-controlled SI. \textbf{Hence, our method must rely less on content information from $\contentd$ while retaining the author style information from $\styled$.} As $\contentd$ and $\styled$ are latent, our method does not estimate them directly but instead adopts a geometric intervention that reduces content-oriented components to achieve our objective (\cref{sec:geometric-interventions}).

\subsection{Related Works} \label{sec:related-works}

\textbf{Authorship attribution (AA).} 
Existing AA methods typically either:
\textbf{(a)} rely primarily on content information, and \textbf{(b)} attempt to reduce content information.
Works in \textbf{(a)} (e.g., BertAA \citep{fabien-etal-2020-bertaa}, STAR \citep{huertastato2023understandingwritingstylesocial}, PART \citep{huertastato2025partpretrainedauthorshiprepresentation} and TRACE \citep{wang2024tracetransformerbasedattributionusing}) treat each text as a single representation, often cause content signals to dominate and may exploit semantic content shortcuts. 
Recent LLM-based AA methods use in-context learning~\citep{huang-etal-2024-large} where reference texts for all candidate authors are provided in prompts, but face challenges in scaling due to context-window limits and high inference cost, and do not generalize well across domains. 
In contrast, works in \textbf{(b)} explicitly emphasize style and attempt to reduce content dependence. 
\textit{Stylometry}\citep{gomez2018stylometry} directly extracts numeric stylistic features to train classification models, but its performance declines quickly with number of authors. Other works include those training content-independent style embeddings by contrastive learning \citep{wegmann-etal-2022-author, patel-etal-2025-styledistance} and separate style and content encoders by minimizing their mutual information \citep{wegmann-etal-2022-author, patel-etal-2025-styledistance}.
However, these methods tend to remove semantic information and yield low performance in real AA tasks. Recent works such as Waterfall \citep{waterfall} proposed an active AA approach where text are paraphrased with LLMs to embed author-specific watermarks. Such watermarks are robust and can be used to track LLM data provenance \cite{watershed, waterdrum}, but require active data processing that may impact the author's original style. EDITOR presents a complementary approach for attribution without edits to the original text. 

\textbf{Style imitation (SI).} Similarly, existing SI methods typically either: \textbf{(a)} do not attempt to separate content and style, and \textbf{(b)} focus on learning style by controlling or removing content. For \textbf{(a)}, \citep{kar94992, liu-etal-2024-customizing-large, lau-etal-2018-deep} directly fine-tune a language model on a target author corpus but risk memorizing training passages. Prompt-based methods \citep{cho-etal-2025-tuning, wang-etal-2025-catch} are tuning-free and flexible, but their stylistic control is limited by context-window size and prompt sensitivity \citep{arora2023ask, laudipper}, tending to capture surface markers rather than implicit writing habits. For \textbf{(b)},
early methods \citep{jhamtani-etal-2017-shakespearizing} use true parallel source-target pairs which are rare and expensive to construct for real authors. Recent methods such as ASTRAPOP \citep{liu2024authorshipstyletransferpolicy} relax this constraint by constructing pseudo-parallel data by neutral paraphrasing or back-translation, then training models against style, content, and fluency rewards. However, most RL-based methods operate on coarse attributes such as sentiment, genre or formality \citep{luo2019dualreinforcementlearningframework, hallinan-etal-2023-steer, liu-may-2025-style} rather than individual author style. Methods such as StyleLM \citep{syed2020stylelm}, StoryTrans \citep{zhu-etal-2023-storytrans} and ParChoice \citep{grondahl2020effectivewritingstyleimitation} rewrite arbitrary input text in a target style through denoising autoencoders, combinatorial paraphrasing etc. However, they are typically evaluated on a few authors with highly distinctive voices and only operate on paragraph-level rewriting rather than long-passage generation from compact content specification. While not applied to SI, past works have studied parameter-space subspace projection for model editing and LoRA fusion of separate tasks \citep{wang2023orthogonal,fang2025alphaedit,chen2026nploranullspaceprojection}. Unlike these works, \EDITORS applies activation-space deflation that allows it to disentangle style and content within text (see \cref{sec:related_work_deflation} for details).

Unlike existing works, our method \EDITORS learn style while optionally retaining content information with our diagnose-then-deflate approach, significantly improving AA and SI performance over baselines. It does not require specially curated data, scales to larger author sets, and can generate long stylized passages from compact content specifications.

\section{The \EDITORS Framework}
\label{sec:method}

\begin{figure}[t]
  \centering
  \includegraphics[width=.9\linewidth]{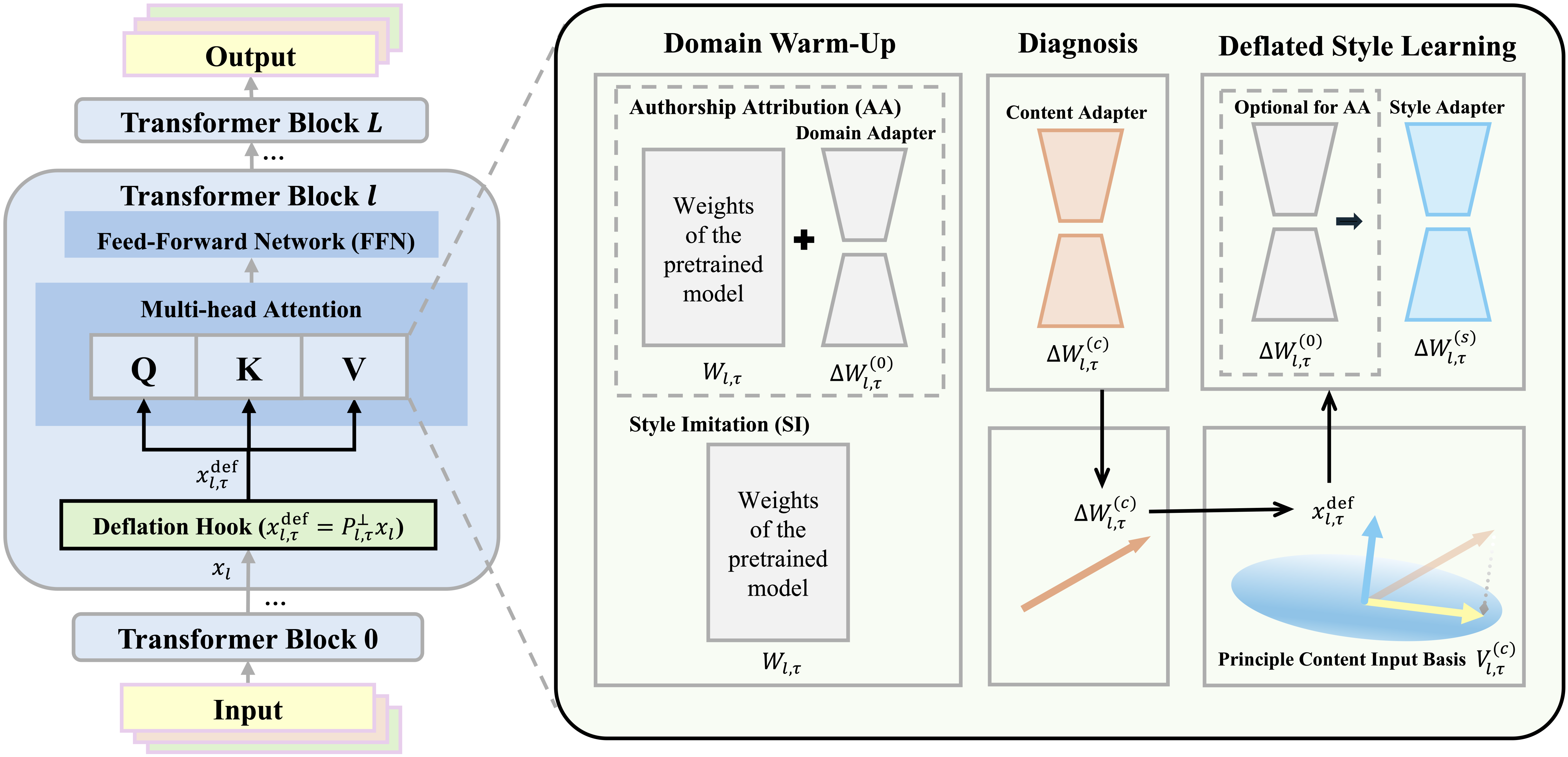}
  \vspace{-1mm}
  \caption{Overview of the \EDITORS pipeline. %
  }
  \label{fig:three_stage_pipeline}
  \vspace{-2mm}
\end{figure}

Our training framework \EDITORS adapts a pretrained LLM to perform AA and SI and prioritizes stylistic features. \cref{fig:three_stage_pipeline} gives an overview of the \EDITORS framework and \cref{alg:editors} gives the pseudocode. \EDITORS follows a \emph{diagnose-then-deflate} strategy, where we first train a diagnostic LoRA adapter on style-neutral statements and use it to identify activation directions that are predictive for semantic content. We then deflate the input activations of training data away from these directions and train a new LoRA adapter (i.e., \textit{style adapter}). During inference, the style adapter introduces author-specific style features, while the frozen base model provides
necessary content support.

\subsection{Geometric Interventions in LLM Representations} \label{sec:geometric-interventions}

As described in \cref{sec:problem_formulation}, each author $a$ induces a marginal passage distribution $\parad$ by combining a content distribution $\contentd$ with an expression
distribution $\styled(\cdot\mid c)$. Ideally, for robust results, AA should avoid relying only on differences in $\contentd$, and (content-controlled) SI should approximate $\styled(\cdot\mid \ddot{c})$ for user-specified content $\ddot{c}$. However, in practice $\contentd$, $\styled$ and the latent content $c$ are not directly observed, and cannot be clearly identified/disentangled from observed text $t \sim \parad$. 

Instead of estimating the distributions directly, \EDITORS makes use of pretrained LLMs' representation space, with its rich text semantic and stylistic features, as an operational domain to intervene on style-content entanglement. Specifically, let $x_{\ell} \in\mathbb{R}^{d_{\mathrm{in}}}$ denote the input activation produced by the frozen pretrained model at layer \(\ell\) (e.g., for passage $t$). Samples from the marginal distribution $\parad$ induce a corresponding distribution over activations \(x_{\ell}\). As discussed in \cref{sec:problem_formulation}, variation in text and consequently the activations can arise from both the latent content \(c\sim\contentd\) and the author's expression pattern \(\styled(\cdot\mid c)\). A standard adapter trained on $\mathcal{D} =\{(t_i,a_i)\}_{i=1}^{N}$ may therefore learn directions in activation space that are predictive of the author only due to content patterns from \(\contentd\), making it less robust to any test-time distribution shift of $\contentd$. 

Our key intuition is that the dominant content-oriented variation in \(x_{\ell}\) can be captured by a low-dimensional linear subspace. Deflating the input activations $x_\ell$ away from this subspace will then give a tractable way to focus training of a style adapter on modeling an author's stylistic features. In \cref{sec:method_diagnosis}, we explain how we can practically identify the diagnostic content-oriented subspace; in \cref{sec:method_deflation}, we describe how we can use the identified subspace to prioritize style learning.

\subsection{Diagnosing Content-Oriented Directions}
\label{sec:method_diagnosis}

Intuitively, the content-oriented directions will dominate in an adapter trained using content-heavy and style-neutral passages. Such passages can be effectively obtained in practice by prompting an LLM to rewrite each training passage $t_i$ into pieces of style-neutral statements\footnote{In contrast, it is infeasible to prompt an external LLM to generate content-controlled passages with different author styles (required by existing works) due to poor SI performance.}.
Using this technique (more details in \cref{app:prompt_atomic}), we first construct an auxiliary style-neutral training corpus $\mathcal{D}^{(c)}$, which contains pieces of style-neutral statements that preserve entities, events,
claims, and other semantic information while suppressing author-specific expression patterns. These statements \(\mathcal{D}^{(c)}\) serve as an
observable proxy for latent content \(c \sim C_a\), and we make a local geometric assumption
that
the directions most useful for a content-oriented auxiliary task ($\mathcal{D}^{(c)}$) provide an approximate basis for content variation in the original task ($\mathcal{D}$). We empirically verify this assumption in \cref{app:content_probe}.

Thus, to identify the content-oriented directions, we start by training a \textit{content adapter} \(\Delta W^{(c)}\) from scratch using the standard next-token prediction objective on
$\mathcal{D}^{(c)}$. For a selected LoRA module \((\ell,\tau)\) at layer $\ell$, the
content-adapted computation is
\(
    h^{(c)}_{\ell,\tau} = \smash{\left(W_{\ell,\tau} + \Delta W^{(c)}_{\ell,\tau}\right)}\ x_\ell\ ,
\)
where \(W_{\ell,\tau}\) is the frozen base model. \textbf{The content adapter is not used in the final
model, but is used only as a diagnostic object.} As it is trained for a
content-oriented task, the input directions to which it is most oriented provide an estimate of content-oriented directions in the pretrained representation space.

To extract the content-oriented directions, we then apply singular value decomposition (SVD) \citep{abdi2007singular} to the content adapter weights 
\(
    \smash{\Delta W^{(c)}_{\ell,\tau} = U_{\ell,\tau} \Sigma_{\ell,\tau} V_{\ell,\tau}^\top} \ 
\)
for every LoRA module $(\ell, \tau)$. Since $\Delta \smash{W_{\ell,\tau}^{(c)}\in \mathbb{R}^{d_{\text{out}} \times d_{\text{in}}}}$ acts on the input activation
$x_\ell$%
, its right singular vectors $V_{\ell,\tau}$ describe directions in the
input activation space. We select the top-\(k\) of them as the
\textit{principal content input basis} :
\begin{equation}
V^{(c)}_{\ell,\tau} \;=\; V_{\ell,\tau} [:, 1:k]\in \mathbb{R}^{d_{\text{in}} \times k}\ .
\label{eq:svd_basis}
\end{equation}

The dimension \(k\) controls the strength of the intervention: larger \(k\)
removes more diagnosed content-oriented directions, while smaller \(k\) retains
more content-oriented directions. Note that this parameter is distinct
from the LoRA rank \(r\), which controls adapter capacity.

\subsection{Deflated Style Learning}
\label{sec:method_deflation}

Given the content-oriented basis \(V_{\ell,\tau}^{(c)}\), we define the %
deflation operator\footnote{More background on deflation can be found in \cref{sec:related_work_deflation}.}
\begin{equation*} \textstyle
    P_{\ell,\tau}^{\perp} = I - V_{\ell,\tau}^{(c)} \left(V_{\ell,\tau}^{(c)}\right)^\top  \in \mathbb{R}^{d_{\text{in}} \times d_{\text{in}}}\ ,
\end{equation*}
which projects input activation $x_\ell$ to the orthogonal complement of \(V_{\ell,\tau}^{(c)}\). The deflated input is $\smash{x_{\ell,\tau}^{\text{def}} = P_{\ell,\tau}^{\perp}\  x_\ell  \in \mathbb{R}^{d_{\text{in}}}}\ $.
By construction, $\smash{(V_{\ell,\tau}^{(c)})^\top x_{\ell,\tau}^{\text{def}}=0}$ and hence the component of the activation lying in the diagnosed content-oriented subspace will be absent from the module input. 
Note that \EDITORS only removes the top-\(k\) directions identified by the content adapter $\smash{\Delta W^{(c)}_{\ell,\tau}}$ at selected modules, instead of all semantic information.

We then train the final \textit{style adapter} \(\Delta W_{\ell,\tau}^{(s)}\) on the
original author-labeled corpus using the deflated input activation $\smash{x_{\ell,\tau}^{\text{def}}}$ at each layer $\ell$ (also used during inference):
\begin{equation} \textstyle
    h_{\ell,\tau}^{(s)} = \left(W_{\ell,\tau} + \Delta W^{(s)}_{\ell,\tau}\right)\ x_{\ell,\tau}^{\text{def}}=
    \left(W_{\ell,\tau} + \Delta W_{\ell,\tau}^{(s)}
    \right)
    P_{\ell,\tau}^{\perp}x_{\ell}\ .
    \label{eq:stage2_deflation}
\end{equation}
\cref{eq:stage2_deflation} applies deflation to the full module input rather than only to the LoRA branch: neither the frozen base weights nor the trainable style update receives the component from the selected content input basis. Note however that the intervention is local and partial: it is applied only to selected attention modules and only removes a \(k\)-dimensional subspace, leaving the remaining activation space and other layers available for linguistic processing.

\textbf{Activation- vs. parameter-space intervention.} \EDITORS intervenes in the input activation space rather than directly constraining the style LoRA weights. Note that the diagnosed basis \(\smash{V_{\ell,\tau}^{(c)}}\) is in the input activation space where content-oriented variation was analyzed, and not the parameter space of the style adapter unlike past works\citep{wang2023orthogonal,fang2025alphaedit,chen2026nploranullspaceprojection}. In contrast, a constraint such as
\(\smash{\Delta W_{\ell,\tau}^{(s)}\perp \Delta W_{\ell,\tau}^{(c)}}\) would only impose a parameter-space relation between two matrices and not ensure that the style adapter avoids content-correlated directions in the actual input
distribution. Deflating activations constrains the signal available during
training, forcing the style adapter to learn from the residual
representation from the start rather than have the style adapter first learn an entangled mixture of content and style before trying to reduce content influence post hoc. We empirically justify this design choice in \cref{app:posthoc}.

\subsection{Task-Specific Training}
\label{sec:method_task_training}

\EDITORS can be applied to both AA and SI tasks. The main \textit{diagnose-then-deflate} procedure above is shared by AA and SI: In both settings, the same deflation operators \(P_{\ell,\tau}^{\perp}\) are used during training. The deployed \EDITORS model consists of the frozen base model $W_0$, the deflation operators \(P^{\perp}\), and only the final style adapter $\Delta W^{(s)}$. The difference lies only in what supervision the model receives and how the final style adapter is initialized. We will describe the details below.

\textbf{Authorship attribution (AA).} The adapter is trained with a classification objective: Given a passage
\(t_i \in \mathcal{T}\), the model is trained to output its author label \(a_i \in \mathcal{A}\). During inference for AA, the model is used to predict an author label from an input passage. 

To initialize the style adapter, we first perform domain/task warm-up by training a separate LoRA \(\Delta W^{(0)}\) on the original training corpus $\mathcal{D}$ for a few steps. We then initialize the style adapter from this domain-adapted adapter:
\(
    \smash{\Delta W_{\ell,\tau}^{(s)} = \Delta W_{\ell,\tau}^{(0)}} \ ,
    \label{eq:AA_init}
\)
and then train it on $\mathcal{D}$ using the deflated forward pass in \cref{sec:method_deflation}. Specifically, the domain warm-up initialization preserves useful task and corpus information (see ablation in \cref{app:domain_warmup,tab:aa_warmup_ablation}) which aligns with broader findings in training data mixture optimization and domain adaption\citep{chen2026duet,xie2023doremi},
while the deflated training reduces access to the dominant
content-oriented directions. 

\textbf{Style imitation (SI).} The adapter is trained on $\mathcal{D}$ with an author-conditioned language modeling objective: given the target author
\(a_i \in \mathcal{A}\), the model is trained to generate the corresponding passage \(t_i \in \mathcal{T}\). During inference for SI, the model receives a target author \(a\) and a content specification \(\ddot{c}\) (for content-controlled SI), and generates a passage conditioned on them.

For SI, we instead initialize the style adapter from zero:
\(
    \Delta W_{\ell,\tau}^{(s)}
    =
    \mathbf{0}.
    \label{eq:SI_init}
\)
There is no domain warm-up initialization like AA because an adapter first trained on the target authors' original passages may encode author-specific topics, entities, or narrative patterns. Such content patterns can increase memorization risk and weaken control by the user-provided content specification $\ddot{c}$, as we show empirically in \cref{sec:style_imitation}. For readers' convenience, we summarize all objects that appear in the \EDITORS pipeline in \cref{tab:adapter_lifecycle}.

\begin{table}[t]
\caption{
Lifecycles of objects in \EDITORS. The content adapter
\(\Delta W^{(c)}\) is discarded after extracting the operator
\(P^{\perp}\). The deployed model uses the frozen base
\(W\), \(P^{\perp}\), and the style adapter \(\Delta W^{(s)}\).
}
\label{tab:adapter_lifecycle}
\centering
\small
\renewcommand{\arraystretch}{0.8}
\begin{tabular}{lllll}
\toprule
Object & Trained on & Purpose & Used at inference? & Initialization \\
\midrule
\(W\) & Pretraining & Frozen base model & Yes & -- \\
\(\Delta W^{(0)}\) & $\mathcal{D}$ (optional) & Domain/task warm-up & No & $\mathbf{0}$ \\
\(\Delta W^{(c)}\) & Style-neutral $\mathcal{D}^{(c)}$ & Diagnose content directions & No & $\mathbf{0}$ \\
\(P^{\perp}\) & Derived from \(\Delta W^{(c)}\) & Deflate content directions & Yes & SVD of \(\Delta W^{(c)}\) \\
\(\Delta W^{(s)}\) & $\mathcal{D}$ (under deflation) & Final style adapter & Yes & Task-dependent \\
\bottomrule
\end{tabular}
\end{table}

\section{Experiments}
\label{sec:experiments}

We evaluate \EDITORS on AA (Sec.~\ref{sec:aa_experiments}) and
SI (Sec.~\ref{sec:style_imitation}). For main LoRA-based experiments, we target the attention query, key, and value projections with rank $r=8$, scaling factor $\alpha=16$, and dropout $0.1$. Full implementation settings, more details and results are in \cref{app:dataset_stats}-\ref{app:ablations}.

\subsection{Authorship Attribution (AA)}
\label{sec:aa_experiments}

\textbf{Datasets \& evaluation metric.}
We evaluate on three closed-set AA benchmarks spanning distinct domains: \textbf{CCAT10/50}~\citep{reuter_50_50_217}, \textbf{Blogs10/50}~\citep{tyo-etal-2023-valla}, and \textbf{Gutenberg30}~\citep{lahiri:2014:SRW}, covering news, blog, and literary domains. %
Further details of datasets are provided in App.~\ref{app:dataset_stats}. We use classification accuracy (i.e., $
\smash{(1/N)
\sum_{i=1}^{N}
\mathbf{1}
\!\left[
\hat{a}_i = a_i
\right]}$) as the evaluation metric (more in App.~\ref{app:aa_metrics}).

\textbf{Base LLMs \& baselines.}
Our main backbone is Llama-3.1-8B-Instruct~\citep{grattafiori2024llama3herdmodels}. To evaluate whether \EDITORS generalizes across architectures, we also test on BERT~\citep{devlin-etal-2019-bert} (without LoRA), Phi-4-mini-Instruct~\citep{microsoft2025phi4minitechnicalreportcompact}, Mistral-7B-Instruct-v0.3~\citep{jiang2023mistral7b}, Gemma2-9B-it~\citep{gemma_2024}, Qwen3.5-9B~\citep{qwen3.5} and Gemma-4-E4B-it~\citep{gemmateam2026gemma4}. We compare against representative baselines from three families: \textit{SFT methods} (\textbf{BertAA}~\citep{fabien-etal-2020-bertaa} and \textbf{STAR}~\citep{huertastato2023understandingwritingstylesocial}), \textit{pure style embedding methods} (\textbf{StyleEmbedding} \citep{wegmann-etal-2022-author} and \textbf{StyleDistance} \citep{patel-etal-2025-styledistance}), and \textit{prompt-based LLM attribution} (\textbf{GPT-4, GPT-5 nano, Qwen3.5-27B~\citep{qwen3.5} + LIP}~\citep{huang-etal-2024-large}). We also include a plain SFT baseline without deflation to isolate the contribution of \EDITORS. Implementation details and inference prompts are provided in App.~\ref{app:aa_implementation} and ~\ref{app:prompt_aa}.

\textbf{Main results.}
We first compare \EDITORS on five benchmarks with AA baselines, along with an additional naïve baseline that we introduced -- an LLM finetuned on the benchmark's respective training data (Llama-3.1-8B SFT).
Each of the benchmarks evaluates methods on different types of challenges associated with AA. CCAT50 is a topic-controlled news benchmark where authors often discuss overlapping corporate events and entities, making coarse topical cues less reliable. 
Blogs50 is noisy and naturally imbalanced, testing robustness to uneven author coverage and informal writing. Gutenberg30 is a long-form literary benchmark where AA depends on discourse-level style cues that are difficult for short-context encoder baselines to capture.

As seen in \cref{tab:aa_results}, \EDITORS achieves the best accuracy on all five benchmarks, outperforming all baselines by a significant margin. Note that our newly introduced SFT baseline (Llama SFT) surprisingly also performs very competitively against baselines, indicating that several of the baseline methods may have specialized approaches that might have disadvantaged them in achieving generalizable performance, unlike \EDITORS which consistently outperforms across benchmarks. \EDITORS's gains are especially large on more challenging 50-author settings, increasing accuracy from $67.33\%$ to $80.40\%$ on CCAT50 and from $65.33\%$ to $83.80\%$ on Blogs50. As the candidate author set grows, authors are more likely to share overlapping topics. \EDITORS's approach of deflating content-oriented directions may have supported its strong performance in such realistic scenarios with more overlapping content and the model must rely more heavily on style cues.

The inconsistent performance of baselines may be indicative of challenges around their method's generalizability. For example, BertAA is competitive on shorter news and blog texts but performs poorly on Gutenberg30, likely because long literary passages exceed the effective input constraints of BERT-style encoders. STAR, which is pretrained on large datasets, does not consistently scale across domains and larger candidate sets. StyleEmbedding and StyleDistance underperform broadly, suggesting that aggressively removing content loses useful author-discriminative information. GPT-4 + LIP is limited to small candidate sets because reference-passage prompting does not scale to 30-50 authors. While our SFT baseline (Llama SFT) is surprisingly robust, it faces challenges in generalizing beyond its training dataset when required to handle larger candidate sets.

\begin{table}[t]
\centering
\caption{AA accuracy (\%). \textbf{Bold} and \underline{underline} mark the best and second-best values. Some results for GPT-4 + LIP are marked with -- because providing necessary reference passages for $\geq30$ authors exceeds its context window.}
\small
\setlength{\tabcolsep}{4pt}
\renewcommand{\arraystretch}{0.8}
\begin{tabular}{lccccc}
\toprule
\textbf{Method} & \textbf{CCAT10} & \textbf{CCAT50} & \textbf{Blogs10} & \textbf{Blogs50} & \textbf{Gutenberg30} \\
\midrule
BertAA%
& 84.00 & 65.72 & 65.40 & 59.69 & 42.76 \\
STAR%
& \underline{91.75} & \underline{67.33} & 73.40 & \underline{65.43} & 53.30 \\
StyleEmbedding%
& 18.20 &  4.28 & 40.67 & 15.60 &  5.00 \\
StyleDistance%
& 20.60 &  8.00 & 52.33 & 15.40 & 11.67 \\
GPT-4 + LIP%
& 57.80 & -- & 68.00 & -- & -- \\
GPT-5-nano + LIP
& 57.00 & -- & 35.18 & -- & -- \\
Qwen3.5-27B + LIP
& 57.00 & -- & 38.16 & -- & -- \\
\midrule
Llama-3.1-8B SFT 
& 86.00 & \underline{67.33} & \underline{93.67} & 65.33 & \underline{75.67} \\
\textbf{EDITORS} 
& \textbf{96.33} & \textbf{80.40} & \textbf{95.67} & \textbf{83.80} & \textbf{82.33} \\
\bottomrule
\end{tabular}
\label{tab:aa_results}
\end{table}

\begin{table}[t]
\centering
\caption{AA accuracy (\%) across backbone architectures. $\Delta_{\text{avg}}$ denotes the average absolute improvement over plain SFT across the five benchmarks; $\sigma_{\Delta}$ is the standard deviation across benchmarks. For BERT, the baseline is BertAA; for LLM backbones, the baseline is plain LoRA SFT.}
\small
\setlength{\tabcolsep}{3.2pt}
\renewcommand{\arraystretch}{0.8}
\begin{tabular}{llccccc|cc}
\toprule
\textbf{Backbone} & \textbf{Setting} 
& \textbf{CCAT10} & \textbf{CCAT50} 
& \textbf{Blogs10} & \textbf{Blogs50} 
& \textbf{Gutenberg30} 
& $\Delta_{\text{avg}}$ 
& $\sigma_{\Delta}$ \\
\midrule
\multirow{2}{*}{BERT} 
& BertAA      & 84.00 & 65.72 & 65.40 & 59.69 & 42.76 & -- & -- \\
& \textbf{EDITORS} & \textbf{90.75} & \textbf{77.47} & \textbf{92.68} & \textbf{83.39} & \textbf{43.57} & +14.1 & 10.0 \\
\midrule
\multirow{2}{*}{Phi-4-mini-3.8B} 
& SFT      & 93.70 & 69.07 & 91.25 & 71.60 & 48.03 & -- & -- \\
& \textbf{EDITORS}  & \textbf{95.70} & \textbf{70.40} & \textbf{92.08} & \textbf{77.74} & \textbf{53.37} & +3.1 & 2.2 \\
\midrule
\multirow{2}{*}{Mistral-7B} 
& SFT      & 93.00 & 80.53 & 94.17 & 75.79 & 72.37 & -- & -- \\
& \textbf{EDITORS}  & \textbf{97.25} & \textbf{86.80} & \textbf{95.77} & \textbf{85.65} & \textbf{73.00} & +4.5 & 3.3 \\
\midrule
\multirow{2}{*}{Llama-3.1-8B} 
& SFT      & 86.00 & 67.33 & 93.67 & 65.33 & 75.67 & -- & -- \\
& \textbf{EDITORS}  & \textbf{96.33} & \textbf{80.40} & \textbf{95.67} & \textbf{83.80} & \textbf{82.33} & +10.1 & 6.3 \\
\midrule
\multirow{2}{*}{Gemma2-9B} 
& SFT      & 85.30 & 53.87 & 64.52 & 50.40 & 61.53 & -- & -- \\
& \textbf{EDITORS}  & \textbf{96.30} & \textbf{74.53} & \textbf{93.27} & \textbf{81.61} & \textbf{65.10} & +19.0 & 10.5 \\
\midrule
\multirow{2}{*}{Qwen3.5-9B}
& SFT
& 52.50 & 74.40 & 93.37 & 73.36 & 54.93
& -- & -- \\
& \textbf{EDITORS}
& \textbf{92.50}
& \textbf{75.00}
& \textbf{95.18}
& \textbf{83.80}
& \textbf{57.83}
& +11.2 & 14.8 \\
\midrule
\multirow{2}{*}{Gemma-4-E4B-it}
& SFT
& 84.33 & 60.80 & 67.27 & 52.26 & 46.33
& -- & -- \\
& \textbf{EDITORS}
& \textbf{92.75}
& \textbf{74.33}
& \textbf{94.05}
& \textbf{72.13}
& \textbf{47.90}
& +14.0 & 8.8 \\
\bottomrule
\end{tabular}
\label{tab:aa_multibackbone}

\end{table}

\textbf{Generalization across base LLMs.}
\cref{tab:aa_multibackbone} shows that \EDITORS consistently improves over our SFT baseline across all backbone LLMs and benchmarks (including the encoder-only BERT backbone). %
Gemma2-9B benefits most with a gain of $+19.0\%$, though with higher variance across datasets ($\sigma_{\Delta}{=}10.5$), %
suggesting the degree of style-content entanglement in Gemma2 varies substantially by domain. Phi4-mini-3.8B, as the most compact LLM base model, shows the smallest and most uniform gain ($+3.1\%$, $\sigma_{\Delta}{=}2.2$). In general for LLM base models, \EDITORS' performance gain over SFT seems to be correlated with model size, though the gains vary fairly widely. This might suggest LLM backbones having different extent of entangled content and style features in their internal representations, and compact models have less redundant capacity for deflation to exploit. Despite these differences, \EDITORS achieves consistent gains, suggesting its generalizability.

\subsection{Style Imitation (SI)}
\label{sec:style_imitation}

We consider two settings from \cref{sec:problem_formulation}: (1) SI, where the model only conditions on the target author $a$, and (2) content-controlled SI, where it must also follow a user-provided content specification $\ddot{c}$. Detailed prompts are provided in App.~\ref{app:prompt_si}.

\begin{wrapfigure}{r}{0.3\textwidth}
\vspace{-5mm}
    \begin{minipage}[t]{0.3\textwidth}
    \centering
    \includegraphics[
        width=0.9\linewidth,
        clip
    ]{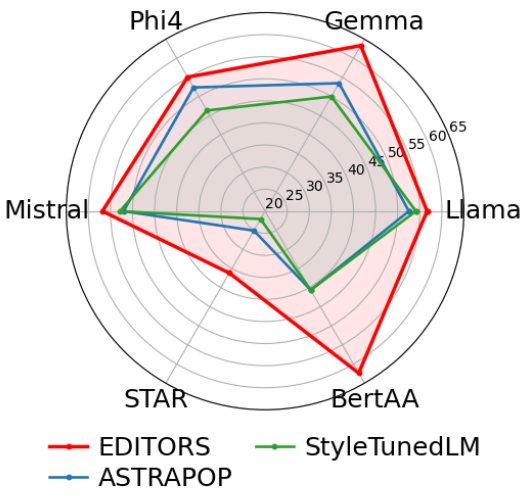}
    
    \caption{All AA evaluators give consistent judgements.}
    \vspace{-3mm}
    \label{fig:si_style}
\end{minipage}
\end{wrapfigure}

\textbf{Dataset \& evaluation metric.}
We focus on \textbf{Gutenberg30} for SI since it exhibits the clearest authorial voice and richest narrative structure among the benchmarks. Specific LoRA configuration can be found in App.~\ref{app:si_implementation}. We evaluate SI along the following axes:
\squishlisttwo
    \item \textbf{Attribution-based style similarity.} We use an ensemble of AA classifiers, including the best performing existing AA methods BertAA and STAR, along with Gemma2-, Llama-, Phi4-, and Mistral-based \EDITORS classifiers (from \cref{sec:aa_experiments}), to measure whether each generation can be attributed to the intended author. %
    We found that while each AA classifier might give different absolute scores, all provide consistent rankings when evaluating SI methods (e.g., see \cref{fig:si_style}).
    \item \textbf{Diversity.} We report BLEU-4~\citep{papineni-etal-2002-bleu, kar94992} against the target author's training corpus, lower BLEU indicates less training-set copying. We also report Self-BLEU, which measures content overlap among same-author generations to assess whether a model repeatedly generates similar author-specific content (lower is better).
    \item \textbf{Preservation of user-provided content} (for (2) content-controlled SI)\textbf{.} We use an ensemble of GPT-4o mini and Qwen3.5-27B to measure content consistency between generated text $\hat{t}$ and user-provided content specification $\ddot{c}$.
\squishend

\vspace{-2.5mm}
App.~\ref{app:si_metrics} provides the detailed metrics, calculation protocols, threshold settings and judge prompts.

\textbf{Base LLMs \& baselines.}
We evaluate \textbf{\EDITORS} with Llama-3.1-8B-Instruct as the base model. %
We compare against four representative SI baselines: \textbf{GPT-4 prompting}~\citep{wang-etal-2025-catch} for tuning-free prompting imitation, \textbf{StyleTunedLM}~\citep{liu-etal-2024-customizing-large} for direct author-corpus fine-tuning, \textbf{ASTRAPOP}~\citep{liu2024authorshipstyletransferpolicy} for RL/preference-based style transfer and \textbf{StoryTrans}~\citep{zhu-etal-2023-storytrans} for non-parallel rewriting. Since ASTRAPOP and StoryTrans require an explicit content input, they are only evaluated in (2) content-controlled SI.

\let\ms\relax

\begin{table}[t]
\centering
\caption{SI results on Gutenberg30. 
Higher is better for style similarity and lower is better for BLEU and Self-BLEU.}
\small
\setlength{\tabcolsep}{5pt}
\renewcommand{\arraystretch}{0.8}
\newcommand{\ms}[2]{#1{\scriptsize$\pm$#2}}
\newcommand{\bms}[2]{\textbf{#1}{\scriptsize$\pm$#2}}
\newcommand{\ums}[2]{\underline{#1}{\scriptsize$\pm$#2}}
\begin{tabular}{lccc}
\toprule
\textbf{Model}
& \textbf{Style Similarity}
& \textbf{BLEU}
& \textbf{Self-BLEU} \\
\midrule

GPT-4
& \ms{23.93}{8.55}
& \bms{4.42}{1.20}
& \ums{27.35}{3.49} \\

StyleTunedLM
& \ums{53.39}{9.43}
& \ms{40.38}{5.42}
& \ms{37.48}{5.60} \\
\midrule

\textbf{\EDITORS} %
& \bms{56.50}{8.74}
& \ums{27.09}{4.76}
& \bms{24.35}{4.90} \\

\bottomrule
\end{tabular}
\label{tab:si_compact_results}
\end{table}

\begin{table}[t]
    \caption{Content-controlled SI results on Gutenberg30. \textit{Content preservation} refers to how much the generated passages adhere to the user-provided content specifications $\ddot{c}$. Higher is better for style similarity and content preservation, while lower BLEU and Self-BLEU indicate less overlap with the training corpus and greater output diversity, respectively. 
    }
\label{fig:si_results_table}
    \centering
    \small
    \setlength{\tabcolsep}{2.5pt}
    \renewcommand{\arraystretch}{0.8}
    \newcommand{\ms}[2]{#1{\scriptsize$\pm$#2}}
    \newcommand{\bms}[2]{\textbf{#1}{\scriptsize$\pm$#2}}
    \newcommand{\ums}[2]{\underline{#1}{\scriptsize$\pm$#2}}
    \begin{tabular}{lcccc}
    \toprule
    \textbf{Model} 
    & \textbf{Style Similarity} 
    & \textbf{Content Preservation} 
    & \textbf{BLEU} 
    & \textbf{Self-BLEU} \\
    \midrule
    GPT-4 
    & \ms{10.82}{0.47}
    & \bms{99.88}{0.06} 
    & \ums{16.54}{0.12}
    & \bms{11.03}{0.13} \\

    StoryTrans 
    & \ms{13.44}{0.52}
    & \ms{0.68}{1.12} 
    & \bms{4.96}{0.95}
    & \ms{14.36}{0.38} \\

    ASTRAPOP 
    & \ums{46.03}{0.56}
    & \ms{3.73}{0.49}
    & \ms{27.08}{0.30}
    & \ms{14.85}{0.50} \\

    StyleTunedLM 
    & \ms{44.34}{0.97}
    & \ms{59.63}{1.34} 
    & \ms{47.18}{0.16}
    & \ms{20.54}{0.39} \\

    \midrule

    \textbf{\EDITORS} %
    & \bms{55.08}{0.56}
    & \ums{68.16}{1.27} 
    & \ms{27.42}{0.64}
    & \ums{13.14}{0.55} \\
    \bottomrule
    \end{tabular}
    \vspace{-1mm}
\end{table}

\textbf{SI.} Tab.~\ref{tab:si_compact_results} shows the results for (1) SI without explicit content constraints. \EDITORS achieves the highest style similarity while substantially reducing BLEU and Self-BLEU compared with StyleTunedLM, showing a better balance between style recognizability, memorization risk, and generation diversity. %

\textbf{Content-controlled SI.}
Tab.~\ref{fig:si_results_table} further evaluates whether models can imitate author styles well while preserving a user-provided content specification \(\ddot{c}\) in (2) content-controlled SI.
\EDITORS achieves the highest style similarity (consistent ranking across all the six evaluators, see \cref{fig:si_style}) while GPT-4 and StoryTrans fail this primary goal. \EDITORS preserves user-provided content specification $\ddot{c}$ (while ASTRAPOP does not), copies less from training corpus (while StyleTunedLM has a larger overlap with it/BLEU), while also achieving better diversity (lower Self-BLEU). 
More comprehensive results from independent style evaluators, additional content preservation assessments, character-level overlap audits, and SI evaluation on Blogs10 are provided in~\cref{app:main_result} (see~\cref{tab:independent_style_evaluation,tab:content_evaluation_robustness,tab:memorization_audit,tab:si_blogs10}). All results demonstrates \EDITORS' effectiveness in achieving balance among style similarity, content preservation and memorization of training content.

\textbf{Qualitative analysis.}
\cref{fig:si_qualitative_comparison} compares generations under the same target author and content specification. \EDITORS preserves the survival content condition, receiving a content score of $5/5$, while all six AA evaluators recognize its generation as Hamlin Garland. The passage expresses Garland-like social observation and plain first-person narration within a survival setting. StyleTunedLM repeatedly uses ``we were together,'' and only two of six  evaluators recognize its generation as the target author.
ASTRAPOP develops a dialogue unrelated to the specified
survival scenario and receives a content score of $1/5$.
\begin{figure}[t]
\centering
\setlength{\fboxsep}{6pt}
\fbox{%
\begin{minipage}{\dimexpr\linewidth-2\fboxsep-2\fboxrule\relax}
\small
\textbf{Target author:} Hamlin Garland.\quad \textbf{Content:} A group works together to survive after
getting stranded.

\medskip
\textbf{StyleTunedLM} (style votes: $2/6$)\par
\smallskip
``We were happy. We were together.
The rain was beating upon us, but we were warm.
The night was dark and cold, but we were together.
The fire was almost out, but we were together.
I was cold, but I was not unhappy. \ldots{} 
We were all that was left of our world.
I reached up and grasped Vi's hand, and she did not pull away.
\ldots{}''

\medskip\hrule\medskip
\textbf{ASTRAPOP} (content: $1/5$)\par
\smallskip
``Half a million people will hear of it,'' continued
Mrs.~Lambert. ``The whole country will ring with it.''
``But she has not consented---'' began Serviss,
only to have Clarke cut him off with a gesture of contempt.
``She will consent when she sees how the thing is coming out.
We have her in the hollow of our hand.  
\ldots{}''

\medskip\hrule\medskip
\textbf{\EDITORS} (style votes: $6/6$; content: $5/5$)\par
\smallskip
``We were all in this venture together, and every man was
willing to risk his life for the sake of the others \ldots{} 
In this matter the two Canadians were very shrewd and took
a leading part in the discussion, while the three of us who
were Americans were content to take the back seats and let
them do the talking. \ldots{} 
At this time I was a tramp,
and I had no home to go to, but I was a tramp who had a home.''
\end{minipage}%
}
\caption{Content-controlled SI generation excerpts.
Style votes count AA evaluators predicting the target
author; content is rated on a $1$--$5$ scale.}
\label{fig:si_qualitative_comparison}
\end{figure}

\subsection{Ablation Studies} \label{sec:experiment-other-results}

\begin{table}[t]
    \caption{Comparison of \EDITORS with and without domain warm-up $\Delta W^{(0)}$ in content-controlled SI. Refer to \cref{tab:si_style_similarity_breakdown} for the comparison in general SI task.
    }
\label{fig:si_w/_w/o}
    \centering
    \small
    \setlength{\tabcolsep}{2.5pt}
    \renewcommand{\arraystretch}{0.8}
    \newcommand{\ms}[2]{#1{\scriptsize$\pm$#2}}
    \newcommand{\bms}[2]{\textbf{#1}{\scriptsize$\pm$#2}}
    \newcommand{\ums}[2]{\underline{#1}{\scriptsize$\pm$#2}}
    \begin{tabular}{lccc}
    \toprule
    \textbf{Model} 
    & \textbf{Style Similarity} 
    & \textbf{Content Preservation} 
    & \textbf{BLEU} \\
    \midrule
    \textbf{\EDITORS} w/ $\Delta W^{(0)}$ 
    & \ms{54.10}{0.68}
    & \ms{39.23}{2.54} 
    & \ms{45.64}{0.26} \\

    \textbf{\EDITORS} w/o $\Delta W^{(0)}$ 
    & \bms{55.08}{0.56}
    & \bms{68.16}{1.27} 
    & \bms{27.42}{0.64} \\
    \bottomrule
    \end{tabular}
\end{table}

In \cref{sec:method_task_training}, we describe an optional domain warm-up that trains an adapter $\Delta W^{(0)}$ to initialize the style adapter $\Delta W^{(s)}$. This warm-up is removed for SI since it may cause the model to memorize training content or disregard user-provided content, as empirically validated in  \cref{fig:si_w/_w/o} (for content-controlled SI) and \cref{tab:si_style_similarity_breakdown} (for general SI). With and without $\Delta W^{(0)}$, \EDITORS achieves similar style similarity, suggesting that $\Delta W^{(0)}$ is not pivotal for inducing style. 
However, 
initializing from $\Delta W^{(0)}$ lowers content preservation by ${\sim} 30$ and increases BLEU by ${\sim} 20$. This suggests that marginal improvements in style similarity for \EDITORS w/ $\Delta W^{(0)}$ may be due more to reproducing author's training passages than successful %
style transfer. In contrast, \EDITORS\ w/o $\Delta W^{(0)}$, our design choice for SI, preserves the user-provided content more faithfully while maintaining comparable style similarity and reducing training-corpus overlap. 

\begin{figure}[htbp]
\centering
\vspace{-0.6em}
\begin{tikzpicture}
\begin{groupplot}[
    group style={
        group size=3 by 1,
        horizontal sep=0.75cm,
    },
    width=0.31\linewidth,
    height=3.0cm,
    symbolic x coords={4,8,16},
    xtick=data,
    grid=major,
    grid style={dashed,gray!25},
    tick label style={font=\scriptsize},
    label style={font=\scriptsize},
    title style={font=\scriptsize, yshift=-0.4em},
    every axis plot/.append style={
        thick,
        mark=*,
        mark size=1.5pt,
    },
]

\nextgroupplot[
    title={Style Similarity $\uparrow$},
    ylabel={Score},
    ymin=46, ymax=51,
    ytick={46,48,50},
]
\addplot[color=blue!70!black] coordinates {
    (4,47.71)
    (8,49.05)
    (16,49.52)
};

\nextgroupplot[
    title={Content Preservation $\uparrow$},
    ymin=50, ymax=72,
    ytick={55,60,65,70},
]
\addplot[color=orange!85!black] coordinates {
    (4,54.68)
    (8,60.75)
    (16,69.50)
};

\nextgroupplot[
    title={BLEU-4 $\downarrow$},
    ymin=24, ymax=31,
    ytick={25,27,29,31},
]
\addplot[color=green!55!black] coordinates {
    (4,29.40)
    (8,26.25)
    (16,25.14)
};

\end{groupplot}

\node[font=\scriptsize, anchor=north] 
at ($(group c2r1.south)+(0,-0.35cm)$)
{Number of deflated content directions $k$};

\end{tikzpicture}
\caption{Sensitivity to the selected content-basis dimension $k$ for SI. Increasing $k$ improves content preservation and reduces BLEU against the training corpus, while style similarity is relatively stable.}
\label{fig:si_k_ablation}
\vspace{-1.5mm}
\end{figure}

In \cref{fig:si_k_ablation}, we investigate how 
the retained content-basis dimension \(k\) (\cref{eq:svd_basis}) affects SI. As \(k\) grows from $4$ to $16$, content preservation increases from $54.68$ to $69.50$, while BLEU against the target author's training corpus decreases from $29.40$ to $25.14$. This suggests that removing a larger portion of the diagnosed content subspace helps the model avoid reproducing training-passage content and better adapt to the user-provided content specification $\ddot{c}$. Style similarity changes only modestly across \(k\), but the higher BLEU at smaller \(k\) indicates that part of the apparent stylistic recognizability may come from training-corpus overlap rather than successful style transfer. In contrast, \(k=16\) achieves the best content preservation and lowest BLEU while maintaining comparable style similarity, suggesting a better balance between style imitation and content control. App.~\ref{app:effect-of-content-basis-dimension} shows a similar effect for AA.

\section{Conclusion}
\label{sec:conclusion}

Existing methods often entangle author style with semantic content, causing AA to rely on content, SI to reproduce training passages, and content-controlled SI to ignore the user-specified content. Our proposed method \EDITORS successfully prioritizes stylistic features without discarding semantic information aggressively by diagnosing a content-oriented subspace and learning style under activation-space deflation, so that selected content-oriented directions are suppressed during training. %
We demonstrate that \EDITORS is highly effective on various AA and (content-controlled) SI benchmarks.
Our results show that deflation behaves as intended: it suppresses content without destroying general semantic information, and activation-space deflation outperforms parameter-space deflation.

\textbf{Limitation and future work.}
Due to the lack of real-world datasets where the same content is expressed in different author styles, we did not provide a more direct ablation controlling for style-specific attribution. We also did not consider uncertainty quantification \citep{kuhn2023semantic,lau2026umpire} of our attributions, which would be valuable in practical deployment.
Although we take the ensemble of classifiers to reduce the SI evaluator bias, the evaluation still relies on automatic evaluators which may not fully capture the real author style unlike human expert evaluations. Future work should therefore focus on building stronger benchmarks for AA with better control over semantic content, and on developing more direct evaluations of how much attribution models rely on style and content respectively.

\section*{Acknowledgments}
This research/project is supported by the National Research Foundation, Singapore under its AI Singapore Programme (AISG Award No: AISG2-PhD/2023-01-039J) and is part of the programme DesCartes. DesCartes: this research is supported by the National Research Foundation, Prime Minister’s Office, Singapore under its Campus for Research Excellence and Technological Enterprise (CREATE) programme. 
This project is partially funded by Microsoft Research Asia.
Xiao Tian and Jue Fan are supported by Agency for Science, Technology and Research
(A*STAR) Graduate Academy. 

\newpage
{\small
\bibliographystyle{plainnat}
\bibliography{references}
}

\newpage
\appendix

\section{Ethical Considerations}\label{sec:ethical-considerations}
AA and SI may be misused for deanonymization, surveillance, unsupported authorship accusations, or impersonation. Our results should therefore be interpreted as model-based evidence rather than definitive authorship proof. Practical deployment should consider uncertainty quantification of attributions to support more holistic investigation. For SI, generated text should be clearly disclosed as machine-generated. We encourage future deployment to include appropriate safeguards, such as watermarking generated text and restricting imitation of living authors without consent.

\section{Algorithm}
\label{sec:algorithm}

\cref{alg:editors} summarizes the \EDITORS training procedure.

\begin{algorithm}[h]
\caption{EDITORS training.}
\label{alg:editors}
\begin{algorithmic}[1]
\Require Passage-author corpus \(\mathcal{D}=\{(t_i,a_i)\}_{i=1}^{N}\);
frozen pretrained model \(W_0\); target modules \(\mathcal{S}\);
LoRA rank \(r\); content-subspace dimension \(k\).
\Ensure Final style adapter \(\Delta W^{(s)}\) and deflation operators
\(P^{\perp}\).

\If{task is AA}
    \State Train domain warm-up adapter \(\Delta W^{(0)}\) on \(\mathcal{D}\).
\EndIf

\State Extract style-neutral statements from passages in \(\mathcal{D}\) to construct $\mathcal{D}^{(c)}$.
\State Train diagnostic content adapter \(\Delta W^{(c)}\) on
$\mathcal{D}^{(c)}$.

\For{each layer \(\ell\) and module \(\tau\in\mathcal{S}\)}
    \State Compute SVD:
    \(\Delta W_{\ell,\tau}^{(c)}
    =
    U_{\ell,\tau}\Sigma_{\ell,\tau}V_{\ell,\tau}^{\top}\).
    \State Set
    \(V_{\ell,\tau}^{(c)}=V_{\ell,\tau}[:,1:k]\).
    \State Construct
    \(P_{\ell,\tau}^{\perp}
    =
    I-
    V_{\ell,\tau}^{(c)}
    (V_{\ell,\tau}^{(c)})^{\top}\).
\EndFor

\If{task is AA}
    \State Initialize
    \(\Delta W^{(s)} \leftarrow \Delta W^{(0)}\).
    \State Train style adapter \(\Delta W^{(s)}\) on the AA objective using deflated
    activations \(x_{\ell,\tau}^{\mathrm{def}}
    =
    P_{\ell,\tau}^{\perp}x_{\ell}\).
\ElsIf{task is SI}
    \State Initialize \(\Delta W^{(s)} \gets \mathbf{0}\).
    \State Train \(\Delta W^{(s)}\) on the author-conditioned LM objective using
    deflated activations
    \(x_{\ell,\tau}^{\mathrm{def}}
    =
    P_{\ell,\tau}^{\perp}x_{\ell}\).
\EndIf

\State \Return \(\Delta W^{(s)}\) and
\(P^{\perp}\).
\end{algorithmic}
\end{algorithm}

\section{Preliminaries}

\subsection{Low-Rank Adaptation (LoRA)} \label{app:lora}

Fully finetuning the layer weights $W_{\ell}$ for every layer $\ell$ of an LLM could be computationally infeasible.
To address this, LoRA \citep{hu2022lora} freezes the pretrained weights $W_{\ell}$ and instead injects a smaller-scale trainable rank-$r$ update. The modified weights at layer $\ell$ become:
\begin{equation*}
W_\ell' = W_\ell + \Delta W_\ell,
\quad\text{with}\
\Delta W_\ell = B_\ell A_\ell,
\label{eq:lora}
\end{equation*}
where $A_\ell \in \mathbb{R}^{r_\ell \times d_{\text{in}}}$,
$B_\ell \in \mathbb{R}^{d_{\text{out}} \times r_\ell}$ are trainable and $r_\ell \ll \min(d_{\text{in}},d_{\text{out}})$. 
The resulting forward computation at layer $\ell$ is
$h_\ell = W'_\ell x_\ell \in \mathbb{R}^{d_{\text{out}}}$, where $x_\ell$ is the input activation at layer $\ell$.

In this paper, we specifically target the multi-head attention (MHA)
projections. For a given block $\ell$, we focus on the Query, Key, and Value modules and denote $\mathcal{S} = \{q, k, v\}$ as the set of target module types.

\subsection{Deflation} \label{sec:related_work_deflation}

\textit{Deflation} (a.k.a. \textit{orthogonal projection}) is a well-known geometric mechanism to decouple representations and mitigate interference between competing signals. It works by projecting a vector to the closest point in a subspace (i.e., the representation of interest) by removing the component orthogonal to that subspace (i.e., the competing signal). %
For example, in activation-space deflation, orthogonal projection 
can efficiently remove information such as gender and topic from embeddings \citep{ravfogel-etal-2020-null}. \citet{qiu2024spectraleditingactivationslarge} edits LLM at inference time by orthogonal projection at hidden activation space to reduce its undesirable behaviors. In the LoRA context, this technique is utilized in continual learning and knowledge editing by operating directly on the weight matrix. Approaches like O-LoRA \citep{wang2023orthogonal}, O-Edit \citep{cai2024oeditorthogonalsubspaceediting} and AlphaEdit \citep{fang2025alphaedit} constrain LoRA updates ($\Delta W_\ell$) to be orthogonal to the subspaces of previously learned tasks or preserved knowledge, thereby preventing catastrophic forgetting. Similarly, NP-LoRA \citep{chen2026nploranullspaceprojection} prevents interference in LoRA fusion by projecting the weights of a content LoRA onto the null space of a style LoRA’s principal directions. 

Motivated by these ideas, we aim to use orthogonal projection to reduce interference between content information and stylistic features in texts. %
Unlike prior weight-centric methods that risk aggressively removing the semantic information, we propose an activation-space deflation mechanism to blind the style adapter to selected semantic signals without directly removing the semantic information. %

\section{Benefits of LoRA to \EDITORS} \label{app:efficient_svd_for_lora}

In \cref{app:lora}, we have introduced the benefits of LoRA to general LLM finetuning. In this section, we will explain why using LoRA can simplify the computation involved in \EDITORS. We also point out that \EDITORS can work with any other SFT methods besides LoRA.

\textbf{LoRA simplifies singular value decomposition (SVD).}

Recall from \cref{sec:method_diagnosis} that we need to perform SVD on the content LoRA adapter $\Delta W^{(c)}$ to extract the content-oriented basis. For each content LoRA adapter we obtained, a direct SVD of
\(
\Delta W_{\ell,\tau}^{(c)}
=
B_{\ell,\tau}^{(c)}A_{\ell,\tau}^{(c)}
\in \mathbb{R}^{d_{\mathrm{out}}\times d_{\mathrm{in}}}
\)
would require explicitly forming a dense matrix of the same size as the original linear layer. This is unnecessary as the update is low-rank. We therefore exploit the LoRA factorization to recover the input-side singular subspace from small matrices whose dimension is at most the LoRA rank \(r\).

For simplicity, we omit the layer and module indices \((\ell,\tau)\) and write
\[
\Delta W^{(c)} = B^{(c)}A^{(c)}, \qquad
B^{(c)}\in\mathbb{R}^{d_{\mathrm{out}}\times r}, \quad
A^{(c)}\in\mathbb{R}^{r\times d_{\mathrm{in}}}.
\]
The content-oriented basis used in Sec.~\ref{sec:method_diagnosis} is the input-side right-singular subspace of \(\Delta W^{(c)}\). Since some rank directions in \(A^{(c)}\) may be annihilated by \(B^{(c)}\), we first remove the null directions of \(B^{(c)}\).
Let $Q_B \in \mathbb{R}^{r \times \rho}$ be an orthonormal basis of $(B^{(c)})^\top$,
where \(\rho=\operatorname{rank}(B^{(c)})\le r\). We then define
\[
A_{\mathrm{eff}} = Q_B^\top A^{(c)}
\in\mathbb{R}^{\rho\times d_{\mathrm{in}}},
\qquad
C = B^{(c)}Q_B
\in\mathbb{R}^{d_{\mathrm{out}}\times \rho}.
\]
Thus,
\[
\Delta W^{(c)}
=
B^{(c)}A^{(c)}
=
C A_{\mathrm{eff}}.
\]
Moreover, \(C\) has full column rank, so
\[
\operatorname{Row}\!\left(\Delta W^{(c)}\right)
=
\operatorname{Row}\!\left(A_{\mathrm{eff}}\right),
\]
where \(\operatorname{Row}(M)\) denotes the subspace spanned by the rows of a matrix \(M\). Therefore, the nonzero right-singular subspace of \(\Delta W^{(c)}\) can be
computed from the much smaller effective factor \(A_{\mathrm{eff}}\).

To obtain the principal content input basis $V^{(c)}_{\ell,\tau}$ in \cref{eq:svd_basis}, we apply SVD to the low-rank effective factor rather than to the full dense update matrix:
\[
A_{\mathrm{eff}}^\top
=
U_A \Sigma_A Z_A^\top,
\qquad
U_A\in\mathbb{R}^{d_{\mathrm{in}}\times \rho},
\quad
\Sigma_A\in\mathbb{R}^{\rho\times \rho},
\quad
Z_A\in\mathbb{R}^{\rho\times \rho}.
\]
The columns of \(U_A\) form an orthonormal basis for the row space of
\(A_{\mathrm{eff}}\). Since \(\Delta W^{(c)}=CA_{\mathrm{eff}}\) and \(C\) has
full column rank, this is the same input-side subspace spanned by the rows of
\(\Delta W^{(c)}\). Therefore, we take
\[
V^{(c)}_{\ell,\tau}
=
U_A[:,1:k]
\in
\mathbb{R}^{d_{\mathrm{in}}\times k}
\]
as the principal content input basis used for activation-space deflation.

In practice, since \(\rho\le r\) and the LoRA rank \(r\) is small, this avoids
forming the dense matrix
\(\Delta W^{(c)}\in\mathbb{R}^{d_{\mathrm{out}}\times d_{\mathrm{in}}}\).
The expensive operation is reduced to SVD-based orthonormalization of the
low-rank factor \(A_{\mathrm{eff}}^\top\), whose rank is at most \(r\). Our
implementation uses this SVD-based solver by default, with QR-based
orthonormalization as an alternative. When all surviving directions are
retained, the SVD and QR solvers yield equivalent deflation subspaces, because
the projection depends only on the span of \(V^{(c)}_{\ell,\tau}\). The SVD
solver is useful when a truncated basis is desired, since it orders the basis
directions by the singular values of \(A_{\mathrm{eff}}^\top\).

\textbf{\EDITORS can be used without LoRA.} Despite the above benefits of LoRA to \EDITORS, we point out that \EDITORS can be used without LoRA too. In \cref{sec:style_imitation}, we include some experiments that finetune BERT models without using LoRA. \EDITORS still achieves a significantly better performance than its baselines such as BertAA.

\section{Datasets}
\label{app:dataset_stats}

\textbf{Gutenberg30} represents the literature domain and is derived from the Project Gutenberg archive~\cite{lahiri:2014:SRW}. To make the evaluation task more challenging, we select $30$  authors who are relatively difficult to recognize, defined as those with the lowest zero-shot classification accuracy under the raw Llama-3-8B-Instruct model among authors who have at least five books available. Each book is segmented into fixed-length passages of roughly $700$ words.

\textbf{Blogs50} represents informal social media writing and is constructed from the Blog Authorship Corpus~\cite{tyo-etal-2023-valla}. This corpus contains $\num{681288}$ blog posts written by $\num{19320}$ bloggers. We select the $50$ bloggers with the largest amount of text from it . In contrast to the other datasets, Blogs50 retains the natural imbalance of the source material, with passage counts varying substantially across authors.

\textbf{CCAT50} represents the news domain and is built from a subset of Reuters Corpus Volume 1~\cite{reuter_50_50_217}. We restrict the corpus to the corporate/industrial topic category (CCAT), yielding a dataset of $50$ authors with $100$ documents per author, for a total of $\num{5000}$ documents.

For all three datasets, the training, validation, and test sets are split strictly by book title or news/blogs category, so that passages originating from the same book or category never appear in multiple splits, ensuring no content overlap between splits to the largest extent. 

\textbf{CCAT10} and \textbf{Blogs10} are randomly sampled subsets of $10$ authors from CCAT50 and Blogs50 respectively. Their split sizes correspond to all passages belonging to the selected $10$ authors. No further subsampling of passages is performed.

\cref{tab:dataset_stats} summarizes the key statistics of the above datasets.

\begin{table}[H]
\centering
\caption{Statistics of the three datasets, including split sizes, the training distribution over authors, and the average passage length. Train Dist. indicates the number of passages per author in the training split. Length denotes the average word count per passage.}
\small
\resizebox{\textwidth}{!}{%
\begin{tabular}{l l rrr c ccc c c}
\toprule
 & & \multicolumn{3}{c}{\textbf{Split Size (Passages)}} & & \multicolumn{3}{c}{\textbf{Train Dist. (Per Author)}} & & \textbf{Length} \\
\cmidrule{3-5} \cmidrule{7-9} \cmidrule{11-11}
\textbf{Dataset} & \textbf{Domain} & \textbf{Train} & \textbf{Val} & \textbf{Test} & & \textbf{Avg} & \textbf{Min} & \textbf{Max} & & \textbf{Avg (Words)} \\
\midrule
\textbf{Gutenberg30} & Literature & 12,000 & 300 & 3,000 & & 400 & 400 & 400 & & 700 \\
\textbf{Blogs50} & Social Media & 42,874 & 5,346 & 5,346 & & 643 & 424 & 1,967 & & 161 \\
\textbf{CCAT50} & News & 4,000 & 250 & 750 & & 80 & 80 & 80 & & 504 \\
\bottomrule
\end{tabular}%
}
\label{tab:dataset_stats}
\end{table}

\section{Implementation}
\label{app:implementation}

\subsection{Authorship Attribution (AA)}
\label{app:aa_implementation}
All models are loaded with 4-bit NF4 quantization and bfloat16 computation, and trained with gradient checkpointing. We attach LoRA adapters to the query, key, and value projection modules, using rank $r{=}8$, scaling factor $\alpha{=}16$, and dropout $0.1$. The diagnostic content adapter $\Delta W^{(c)}$ is trained on the style-neutral statements corpus (App.~\ref{app:prompt_atomic}) via continued instruction tuning with TRL's \texttt{SFTTrainer}. The final style adapter $\Delta W^{(s)}$ then uses the original passage-author supervision, formatted as an instruction-style prompt that asks the model to identify the author based on writing style.

The diagnostic content adapter $\Delta W^{(c)}$ is trained for $2$ epochs with datasets of style-neutral statements. For CCAT, the domain warm-up $\Delta W^{(0)}$ (described in \cref{sec:method_task_training}) is trained for $2$ epochs and the style adapter $\Delta W^{(s)}$ for $4$ epochs. For Blogs and Gutenberg, the domain warm-up $\Delta W^{(0)}$ is trained for $1$ epoch and the style adapter $\Delta W^{(s)}$ for $2$ epochs. We use a maximum sequence length of $\num{1000}$ for both stages across all datasets. Both stages use learning rate $2\times10^{-4}$, warm-up ratio $0.05$, weight decay $0.01$, per-device batch size $12$, and gradient accumulation $2$. For the plain LoRA SFT baseline, we use the same total number of training epochs as the corresponding diagnose-then-deflate \EDITORS configuration. All backbone LLMs are trained with the same hyperparameters.

We do not report error bars for AA because the evaluation is deterministic: we do not use stochastic sampling/sampling-based decoding, so repeated runs of the same checkpoint on the same split produce identical predictions.

\subsection{Style Imitation (SI)}
\label{app:si_implementation}
We use the same LoRA configuration as in the AA experiments. The main difference lies in the supervision format. The diagnostic content adapter $\Delta W^{(c)}$ is reused from the AA setting, and the deflated stage then performs style fine-tuning under activation-space deflation. We evaluate two variants of our method. The default setting skips domain warm-up and trains the style adapter $\Delta W^{(s)}$ directly for $8$ epochs. The second variant first applies domain warm-up for $4$ epochs and then performs deflated style learning for another $4$ epochs, so that the total training budget remains comparable across settings.

At inference time, each of the $30$ Gutenberg authors is paired with the same seven stylistically neutral content summaries listed in App.~\ref{app:prompt_si}. We generate one passage for each author on the given summary, yielding $30\times7=210$ generations for evaluation.

\section{Prompts} \label{app:prompts}

\subsection{Style-Neutral Statements Extraction} \label{app:prompt_atomic}

We prompt GPT-4o mini to extract style-neutral statements from each passage (to be used in \cref{sec:method_diagnosis}). The extracted facts are used as the continued pretraining corpus for the diagnostic content adapter $\Delta W^{(c)}$.

\begin{tcolorbox}[colback=gray!5, colframe=gray!50]
\small
\texttt{\textbf{System:}}\\
\texttt{You turn passages into neutral, styleless atomic facts. Each fact should stand alone, use plain prose, and stay faithful to the passage.}\\[4pt]
\texttt{\textbf{User:}}\\
\texttt{Rewrite this passage as \{min\_facts\} to \{max\_facts\} standalone atomic factual statements.}\\
\texttt{Capture what the passage claims (opinions become factual claims about the author/source).}\\
\texttt{Keep names and entities unchanged; avoid figurative language.}\\
\texttt{Output ONLY a JSON array of strings, nothing else.}\\
\texttt{PASSAGE: \{passage\}}
\end{tcolorbox}

\noindent An example of the original passage and its corresponding style-neutral statements is shown in Figure~\ref{fig:atomic_example}.

\begin{figure}[H]
\centering
\begin{tcolorbox}[title={\small Original literary passage (Edward Phillips Oppenheim, \textit{Jeanne of the Marshes})}, colback=blue!3, colframe=blue!30, left=4pt, right=4pt]
\small
``\ldots\ ever since those few days down at that tumbledown old place of Cecil de la Borne's, she seems to have developed in a remarkable manner \ldots\ I am sure \ldots\ that we are going to have trouble.'' Forrest replied that, so far as he was concerned, the trouble had already arrived, and admitted that he felt exhausted by the life he had been leading. The Princess accused him of cowardice and suggested that he was no longer the man he had once been. Their conversation then turned to the De la Borne money, to De Brensault's intentions toward Jeanne, and to the possibility that Jeanne's obstinacy would bring consequences.
\end{tcolorbox}

\vspace{4pt}

\begin{tcolorbox}[title={\small Extracted style-neutral statements}, colback=green!3, colframe=green!30, left=4pt, right=4pt]
\small
(1) The Princess noted that Jeanne had developed in a remarkable manner after spending time at Cecil de la Borne's place.\\
(2) Forrest said that trouble had already arrived for him.\\
(3) The Princess suggested that Forrest was not the man he used to be.\\
(4) Forrest admitted that he had bullied the Princess for many years.\\
(5) The Princess asked about the De la Borne money, which Forrest confirmed had been paid by his brother.\\
(6) The Princess said that De Brensault was serious about Jeanne.\\
(7) The Princess indicated that Jeanne would suffer consequences if she remained obstinate.
\end{tcolorbox}
\caption{Example of a literary passage and its corresponding style-neutral statements.}
\label{fig:atomic_example}
\end{figure}

\subsection{Authorship Attribution (AA)}
\label{app:prompt_aa}
For authorship attribution, the same prompt is used during both training and inference. The model is instructed to identify the author based on writing style rather than content, and to return only the exact author name. During training, only the author name tokens contribute to the loss.

\begin{tcolorbox}[colback=gray!5, colframe=gray!50]
\small
\texttt{\textbf{System:}}\\
\texttt{You are an expert literary analyst.}\\
\texttt{Identify the author of the provided passage based on writing style rather than content.}\\
\texttt{Answer with only the exact author name.}\\[4pt]
\texttt{\textbf{User:}}\\
\texttt{Passage: \{passage\}}\\
\texttt{Author:}
\end{tcolorbox}

\subsection{Style imitation (SI) prompts}
\label{app:prompt_si}
During training, the model receives the author name as input and learns to generate the corresponding original passage. At inference time, an additional summary field is provided so that the model can follow user-specified content while imitating the target author's style.

\vspace{0.3em}
\noindent\textit{Training prompt:}
\begin{tcolorbox}[colback=gray!5, colframe=gray!50]
\small
\texttt{\textbf{System:}}\\
\texttt{You are a writing assistant that creates original literary paragraphs.}\\
\texttt{Write a new paragraph in the style of the specified author.}\\
\texttt{Do not copy the source text.}\\[4pt]
\texttt{\textbf{User:}}\\
\texttt{Author: \{author\}}\\
\texttt{Task: Write a new literary paragraph.}
\end{tcolorbox}

\vspace{0.3em}
\noindent\textit{Inference prompt:}
\begin{tcolorbox}[colback=gray!5, colframe=gray!50]
\small
\texttt{\textbf{System:}}\\
\texttt{You are a writing assistant that creates original literary paragraphs.}\\
\texttt{Write a new paragraph in the style of the specified author.}\\
\texttt{Do not copy the source text.}\\
\texttt{If a Summary is provided, reflect its content while following the author's style.}\\[4pt]
\texttt{\textbf{User:}}\\
\texttt{Author: \{author\}}\\
\texttt{Summary: \{content\_pack\}}\\
\texttt{Task: Write a new literary paragraph.}
\end{tcolorbox}

Note that the summary line is omitted in the general SI task.

\textbf{User-provided content specifications.}
To evaluate content-controlled SI, we construct a fixed set of seven short user-provided content specifications \(\ddot{c}\) using GPT-5. These \(\ddot{c}\) are designed to cover broad narrative topics that appear across the Gutenberg30 corpus while remaining compact and stylistically neutral. To make the content specification genre-diverse, we first group the $30$ authors into coarse representative genres based on their dominant literary topics and then design one \(\ddot{c}\) for each genre group. These genre assignments are used only to guide the content specification set construction. During generation, each target author is prompted once with each \(\ddot{c}\) and asked to generate a passage in the author's style. This produces $30 \times 7 = 210$ generations in total. Because every \(\ddot{c}\) is generated in every author's style, many prompts place an author in a topic or genre condition outside the author's most typical domain, making the evaluation more challenging.

\begin{table}[H]
\centering
\caption{Content specifications used for content-controlled SI. Authors are classified to representative genre groups.}
\label{tab:si_content_specs}
\small
\setlength{\tabcolsep}{4pt}
\renewcommand{\arraystretch}{1.12}
\begin{tabular}{c p{0.18\textwidth} p{0.33\textwidth} p{0.37\textwidth}}
\toprule
\textbf{ID} & \textbf{Representative genre} & \textbf{Content specification \(\ddot{c}\)} & \textbf{Authors} \\
\midrule
A & Children / youth moral tale
& A child learns that telling the truth is better than hiding a mistake.
& Charlotte Mary Yonge; Jacob Abbott. \\

B & Adventure / survival / exploration
& A group works together to survive after getting stranded.
& Charles Kingsley; Daniel Defoe; Edward Stratemeyer; George Alfred Henty; Harold Bindloss; Jack London; James Otis; R. M. Ballantyne; Robert Louis Stevenson; Rudyard Kipling. \\

C & Mystery / detective / suspense
& A missing item is found by careful thinking, not force.
& Edward Phillips Oppenheim; G. K. Chesterton; Wilkie Collins. \\

D & Social realism / local color
& A small-town rule causes trouble, and someone speaks up.
& Bret Harte; Hamlin Garland; John Galsworthy; John Morley; William Dean Howells. \\

E & Satire / wit / comic society
& A silly rule is taken too seriously and leads to a mess.
& Frank Richard Stockton; George Bernard Shaw; James Matthew Barrie; Jerome Klapka Jerome; Oscar Wilde; William Makepeace Thackeray; William Wymark Jacobs. \\

F & Speculative science fiction / social thought
& A new technology spreads too fast and causes problems.
& Herbert George Wells. \\

G & Allegory / symbolic reflection
& A person tries to hide something small, but it is noticed anyway.
& Nathaniel Hawthorne; William Butler Yeats. \\
\bottomrule
\end{tabular}
\end{table}

\section{Evaluation Metrics}
\subsection{Authorship Attribution (AA)}
\label{app:aa_metrics}
\paragraph{Classification accuracy.} For AA, we primarily report accuracy, i.e., the proportion of test passages whose predicted author matches the ground-truth label. Formally, for a test set of size $N$, accuracy is
\begin{equation*}
\mathrm{Acc}
=
\frac{1}{N}
\sum_{i=1}^{N}
\mathbf{1}
\!\left[
\hat{a}_i = a_i
\right].
\end{equation*}
Since our setting is closed-set attribution, accuracy directly reflects the practical objective of correctly identifying the author from a fixed candidate set.

\subsection{Style Imitation (SI)}
\label{app:si_metrics}
We evaluate SI along three axes:
\squishlisttwo
\item \textbf{Style similarity.} We use an ensemble of AA classifiers, including Gemma2-, Llama-, Phi4-, and Mistral-based \EDITORS classifiers (from \cref{sec:aa_experiments}), together with BertAA and STAR, to measure whether each generation can be attributed to the intended author. 
    \item \textbf{Memorization.} We report BLEU-4~\citep{papineni-etal-2002-bleu, kar94992} and LCS$_{\max}$~\citep{aerni2025measuring} against the target author's training corpus, Self-BLEU~\citep{zhu2018texygenbenchmarkingplatformtext} among same-author generations.
    Lower BLEU and LCS$_{\max}$ indicate less training-set copying, while lower Self-BLEU indicates greater output diversity.
    \item \textbf{Preservation of user-provided content} (for (2) content-controlled SI)\textbf{.} We use GPT-4o mini to measure content consistency between generated text $\hat{t}$ and user-provided content specification $\ddot{c}$.
\squishend

We include the details for each metric below.

\paragraph{Style similarity.}
To evaluate whether a generated passage reflects the style of the target author, we use an ensemble of six AA classifiers: four \EDITORS-based classifiers (Llama-3.1-8B, Gemma2-9B, Mistral-7B, and Phi4-mini-3.8B), together with STAR and BertAA. Using an ensemble rather than a single classifier reduces evaluator bias and provides a more robust style assessment. Given a generated passage $\hat{p}$ and its intended target author $a$, each classifier independently predicts the author and we compute its accuracy:
\begin{equation*}
\mathrm{Acc}_j
=
\frac{1}{N}
\sum_{i=1}^{N}
\mathbf{1}
\!\left[
\hat{a}_{i,j} = a_i
\right].
\end{equation*}
The final style similarity is the average over all six classifiers.

\paragraph{Content preservation.}
Content preservation is evaluated by the average score of GPT-4o mini and Qwen3.5-27B. For each generated paragraph, the judge receives the target semantic summary and the generated text, and returns a three-way label with an integer score. Concretely, the judge prompt is as illustrated below:

\begin{tcolorbox}[colback=gray!5, colframe=gray!50]
\small
\texttt{You are judging whether the generation reflects the summary to some extent.}\\
\texttt{The generation is long; if it clearly mentions the summary idea anywhere, label it yes.}\\
\texttt{Use `partial' only if it is vaguely related but does not clearly express the summary.}\\
\texttt{Use `no' only if it does not reflect the summary at all.}\\
\texttt{Return JSON only with keys: label, score.}\\
\texttt{label must be one of: yes, partial, no.}\\
\texttt{score must be an integer 1--5 (5 = fully matches summary).}\\[4pt]
\texttt{Summary: \{summary\}}\\
\texttt{Generation: \{generation\}}
\end{tcolorbox}

Let $s_i \in \{1,2,3,4,5\}$ denote the judge score for the $i$-th example. We compute the average judge score over the evaluation set and rescale it to a percentage:
\begin{equation*}
\mathrm{ContentPres}(\%)
=
25\left(\frac{1}{N}\sum_{i=1}^{N} s_i - 1\right).
\end{equation*}
Thus, a higher score indicates that the generated text better preserves the intended semantic content.

\paragraph{Fluency.}
We evaluate fluency with GPT-4o mini. For each generated paragraph, we prompt the judge with the instruction as shown below:

\begin{tcolorbox}[colback=gray!5, colframe=gray!50]
\small
\texttt{Given the text below, give a score between 0 and 100 for the fluency and grammatical correctness of the text.}\\
\texttt{Be strict and only give high scores sparingly.}\\
\texttt{Only output the number and do not output any other text or explanation.}\\[4pt]
\texttt{Text: \{generation\}}
\end{tcolorbox}

The returned value is a scalar score in the range $[0,100]$, where higher scores indicate more grammatically well-formed and coherent generations. Let $f_i$ denote the fluency score assigned to the $i$-th generated paragraph. We report the average fluency over the evaluation set:
\begin{equation*}
\mathrm{Fluency}
=
\frac{1}{N}\sum_{i=1}^{N} f_i.
\end{equation*}
This metric complements style similarity and content preservation by ensuring that the generated text remains readable and natural.

We report BLEU-4~\citep{papineni-etal-2002-bleu, kar94992} and LCS$_{\max}$~\citep{aerni2025measuring} against the target author's training corpus, Self-BLEU~\citep{zhu2018texygenbenchmarkingplatformtext} among same-author generations.
 
\paragraph{BLEU-4.}
For each generated passage $g_i$ with target author $a_i$, we compute a single-hypothesis multi-reference BLEU-4 score against the author's training corpus $\mathcal{R}_{a_i}$. Let $C_n(g_i)$ denote the multiset of $n$-grams in $g_i$, and let $c_g(u)$ be the count of $n$-gram $u$ in $g_i$. For each reference passage $r\in\mathcal{R}_{a_i}$, let $c_r(u)$ be the count of $u$ in $r$. The clipped $n$-gram precision is
\[
p_n(g_i,\mathcal{R}_{a_i})
=
\frac{
\sum_{u\in C_n(g_i)}
\min\left(c_g(u), \max_{r\in\mathcal{R}_{a_i}} c_r(u)\right)
}{
\sum_{u\in C_n(g_i)} c_g(u)
},
\qquad n=1,\ldots,4.
\]
The brevity penalty follows the standard closest-reference convention:
\[
\mathrm{BP}(g_i)=
\begin{cases}
1, & |g_i| > |r_i^\star|,\\
\exp\left(1-\frac{|r_i^\star|}{|g_i|}\right), & |g_i|\le |r_i^\star|,
\end{cases}
\]
where $r_i^\star$ is the reference passage in $\mathcal{R}_{a_i}$ whose length is closest to $|g_i|$, with ties broken by the shorter reference. The per-generation BLEU score is then
\[
\mathrm{BLEU}_{\mathrm{train}}(g_i)
=
100\cdot \mathrm{BP}(g_i)
\cdot
\exp\left(\frac{1}{4}\sum_{n=1}^{4}\log p_n(g_i,\mathcal{R}_{a_i})\right).
\]
We report the mean over all generations:
\[
\mathrm{BLEU}_{\mathrm{train}}
=
\frac{1}{N}\sum_{i=1}^{N}
\mathrm{BLEU}_{\mathrm{train}}(g_i).
\]
Unlike translation evaluation, we use this score as a memorization diagnostic rather than a quality metric: higher values indicate stronger lexical overlap with the target author's training corpus and therefore greater risk of training-content reproduction. Similar corpus-overlap diagnostics have been used to study overfitting and plagiarism in author-style generation~\citep{papineni-etal-2002-bleu, kar94992}.

Because BLEU-4 depends on generation length, corpus size, and reference construction, we interpret it using empirical risk bands calibrated to our Gutenberg30 setting rather than as universal thresholds.

For training-corpus BLEU, we use the following heuristic bands:
\[
\begin{array}{ccl}
\mathrm{BLEU}_{\mathrm{train}} < 20
&:& \text{low corpus overlap},\\
20 \le \mathrm{BLEU}_{\mathrm{train}} < 30
&:& \text{moderate corpus overlap},\\
30 \le \mathrm{BLEU}_{\mathrm{train}} < 40
&:& \text{high corpus overlap},\\
\mathrm{BLEU}_{\mathrm{train}} \ge 40
&:& \text{very high corpus overlap / strong memorization risk}.
\end{array}
\]

\paragraph{Character-level overlap audit.}
We adapt the overlap-based audit of \citet{aerni2025measuring} using a minimum matching span of $25$ characters. For generation $g_i$, let $U_i(t)$ denote the set of character positions covered by contiguous verbatim matches of at least $t$ characters in a reference corpus.
We report the overlap rate as a percentage:
\begin{equation*}
\mathrm{OR}@t
=
\frac{100}{N}
\sum_{i=1}^{N}
\frac{|U_i(t)|}{|g_i|}.
\end{equation*}
We calculate $\mathrm{OR}@25$ against both the target author's training corpus and the global training corpus. We also report the maximum copied span length in characters.

\paragraph{Self-BLEU.}
We use Self-BLEU to measure output diversity. Let $\mathcal{G}_a$ be the set of generated passages for author $a$. For each generation $g_i\in\mathcal{G}_a$, we compute BLEU-4 by treating the remaining generations from the same author as references:
\[
\mathrm{SelfBLEU}(g_i)
=
\mathrm{BLEU}\left(g_i,\mathcal{G}_a\setminus\{g_i\}\right).
\]
The final score averages this leave-one-out BLEU over all authors and generations:
\[
\mathrm{SelfBLEU}
=
\frac{1}{N}
\sum_{a}
\sum_{g_i\in\mathcal{G}_a}
\mathrm{BLEU}\left(g_i,\mathcal{G}_a\setminus\{g_i\}\right).
\]
The BLEU computation uses the same clipped 1-4-gram precision and brevity penalty as above. A lower Self-BLEU indicates more diverse generations, while a higher value suggests that the model repeatedly produces similar lexical patterns or templates. This follows prior text-generation evaluation practice, where Self-BLEU is used to detect mode collapse and lexical repetitiveness~\citep{zhu2018texygenbenchmarkingplatformtext}.

For Self-BLEU, we use the following diversity bands:
\[
\begin{array}{ccl}
\mathrm{SelfBLEU} < 12
&:& \text{high output diversity},\\
12 \le \mathrm{SelfBLEU} < 16
&:& \text{moderate output diversity},\\
16 \le \mathrm{SelfBLEU} < 20
&:& \text{reduced diversity / mild repetition},\\
\mathrm{SelfBLEU} \ge 20
&:& \text{low diversity / template-like repetition risk}.
\end{array}
\]

\paragraph{Longest common substring (LCS$_{\max}$).}
To detect verbatim copying, we compute the longest contiguous word sequence shared between a generated passage and any training passage by the same target author. Let $g_i=(w_1,\ldots,w_m)$ be a generated passage and let $r=(v_1,\ldots,v_\ell)$ be a training passage from $\mathcal{R}_{a_i}$. The word-level longest common substring between $g_i$ and $r$ is
\[
\mathrm{LCS}(g_i,r)
=
\max_{p,q,k}
\left\{
k:
(w_p,\ldots,w_{p+k-1})
=
(v_q,\ldots,v_{q+k-1})
\right\}.
\]
For each generation, we take the maximum over all training passages by the same author:
\[
\mathrm{LCS}(g_i,\mathcal{R}_{a_i})
=
\max_{r\in\mathcal{R}_{a_i}}
\mathrm{LCS}(g_i,r).
\]
We report both the average LCS length and the worst-case maximum:
\[
\mathrm{LCS}_{\mathrm{mean}}
=
\frac{1}{N}\sum_{i=1}^{N}
\mathrm{LCS}(g_i,\mathcal{R}_{a_i}),
\qquad
\mathrm{LCS}_{\max}
=
\max_{1\le i\le N}
\mathrm{LCS}(g_i,\mathcal{R}_{a_i}).
\]
$\mathrm{LCS}_{\max}$ serves as a worst-case memorization indicator: small values usually correspond to short common phrases, whereas very large values indicate passage-level reproduction. In implementation, we build a per-author unigram inverted index over the training passages and greedily extend all candidate word matches to find the maximum contiguous overlap. This follows the exact-overlap principle used in recent work on training-data reproduction and copyright-related memorization risks in LLMs, while our metric reports word-level rather than character-level overlap~\citep{aerni2025measuring}.

\section{Experiments and Ablations}
\label{app:ablations}

\subsection{Effect of Domain Warm-up}
\label{app:domain_warmup}
\paragraph{Authorship Attribution (AA).} We compare \EDITORS with and without domain warm-up for AA. Warm-up initializes the style adapter $\Delta W^{(s)}$ with the domain-adapted weights $\Delta W^{(0)}$ before deflated style learning. Unlike SI, where reproducing author-specific content is undesirable, AA can benefit from lexical, topical, and temporal cues alongside stylistic features. As shown in \cref{tab:aa_warmup_ablation}, warm-up improves AA accuracy across all three datasets, supporting its use for AA.

\begin{table}[H]
\centering
\small
\caption{Effect of domain warm-up on AA accuracy (\%).
Higher is better. \textbf{Bold} indicates the higher accuracy
for each dataset.}
\label{tab:aa_warmup_ablation}
\begin{tabular}{lcc}
\toprule
\textbf{Dataset} & \textbf{\EDITORS} w/o $\Delta W^{(0)}$ & \textbf{\EDITORS} w/ $\Delta W^{(0)}$ \\
\midrule
CCAT50      & 66.80 & \textbf{80.40} \\
Blogs50     & 62.07 & \textbf{83.80} \\
Gutenberg30 & 73.30 & \textbf{82.33} \\
\bottomrule
\end{tabular}
\end{table}

\paragraph{Style imitation (SI).}
As we analyze in \cref{sec:experiment-other-results}, \cref{fig:si_w/_w/o} shows that warm-up yields comparable style similarity but lower content preservation and higher BLEU against the target author's training corpus. We therefore omit domain warm-up for SI and initialize the style adapter from zero.

\subsection{Effect of Content-Basis Dimension} \label{app:effect-of-content-basis-dimension}

\begin{figure}[t]
\centering
\vspace{-0.6em}
\begin{tikzpicture}
\begin{axis}[
    width=0.55\linewidth,
    height=4.0cm,
    ymin=68, ymax=80,
    xmin=3.5, xmax=16.5,
    xtick={4,8,16},
    xlabel={Retained content-basis dimension $k$},
    ylabel={AA Accuracy (\%)},
    grid=major,
    grid style={dashed,gray!25},
    tick label style={font=\scriptsize},
    label style={font=\small},
    legend style={
        font=\scriptsize,
        draw=none,
        fill=none,
        at={(0.02,0.98)},
        anchor=north west
    },
]

\addplot[
    dashed,
    thick,
    gray
] coordinates {
    (3.5,70.67) (16.5,70.67)
};
\addlegendentry{Plain SFT}

\addplot[
    thick,
    mark=*,
] coordinates {
    (4,73.47)
    (8,75.47)
    (16,78.90)
};
\addlegendentry{EDITORS}

\end{axis}
\end{tikzpicture}
\caption{Effect of selected content-basis dimension $k$ on AA accuracy. $k$ controls the number of content-oriented directions used for deflation.}
\label{fig:k_ablation}
\end{figure}

\paragraph{Authorship Attribution (AA).} We further vary the number of content-basis directions $k$ in the diagnostic content adapter $\Delta W^{(c)}$ used for deflation. In this experiment, the content adapter $\Delta W^{(c)}$ uses LoRA rank \(r_c=16\), and we truncate the extracted content basis to \(k\in\{4,8,16\}\). Smaller values of \(k\) remove only part of the content-oriented subspace, whereas \(k=16\) uses the full content basis. As shown in \cref{fig:k_ablation}, attribution accuracy improves as \(k\) increases, with the best result obtained at \(k=16\). This suggests that the most effective attribution model is obtained when the final style adapter $\Delta W^{(s)}$ is trained on directions orthogonal to the full content subspace. In other words, the complete content-oriented basis provides the strongest proxy for content-dominated information, allowing deflation to better suppress semantic shortcuts while preserving the residual author-discriminative signals needed for AA.

\paragraph{Style imitation (SI).}
As we analyze in \cref{sec:experiment-other-results}, \cref{fig:si_k_ablation} similarly shows that increasing the retained content-basis dimension \(k\) improves controllability in SI. As \(k\) grows from 4 to 16, it leads to a better balance between style imitation and content control.

\subsection{General SI Task}
\label{app:si_general_result}
We evaluate the general SI task using a prefix-continuation setting. In this setting, no user-provided content specification \(\ddot{c}\) is given at inference time. Instead, the model receives only the target author and a short five-word prefix, and is asked to continue in the target author's style. We construct a $30 \times 30$ prefix grid over Gutenberg30 authors: for each target author, we generate one continuation from a prefix sampled from that same author and $29$ continuations from prefixes sampled from the other authors, yielding $900$ generations per model. Detailed prompts are provided in App.~\ref{app:prompt_si}. \cref{tab:si_style_similarity_breakdown} gives a more comprehensive version of the results shown in \cref{tab:si_compact_results} in the main paper. 

\begin{table}[H]
\centering
\caption{Style similarity and memorization/diversity results on Gutenberg30. StyleSim Avg. is the mean target-author recognition rate across six style-similarity classifiers. $\sigma_{\mathrm{clf}}$ is the sample standard deviation across these classifier means and reflects evaluator disagreement. BLEU and Self-BLEU are reported as mean $\pm$ standard deviation across authors. \textbf{Bold} and \underline{underline} mark the best and second-best values, respectively; higher is better for style similarity, while lower BLEU and Self-BLEU indicate less training-corpus overlap and greater output diversity, respectively.}
\label{tab:si_style_similarity_breakdown}
\scriptsize
\setlength{\tabcolsep}{3.0pt}
\renewcommand{\arraystretch}{0.8}
\newcommand{\ms}[2]{#1{\tiny$\pm$#2}}
\newcommand{\bms}[2]{\textbf{#1}{\tiny$\pm$#2}}
\newcommand{\ums}[2]{\underline{#1}{\tiny$\pm$#2}}
\begin{tabular}{lcccccccccc}
\toprule
\textbf{Model}
& \textbf{Llama}
& \textbf{Gemma}
& \textbf{Phi4}
& \textbf{Mistral}
& \textbf{STAR}
& \textbf{BertAA}
& \textbf{StyleSim Avg.}
& $\boldsymbol{\sigma_{\mathrm{clf}}}$
& \textbf{BLEU}
& \textbf{Self-BLEU} \\
\midrule
EDITORS w/ $\Delta W^{(0)}$
& \textbf{63.67}
& \textbf{67.89}
& \textbf{60.78}
& \textbf{60.56}
& \underline{39.00}
& \underline{67.00}
& \textbf{59.82}
& 10.64
& \ms{47.17}{8.28}
& \ms{34.83}{6.62} \\

EDITORS w/o $\Delta W^{(0)}$
& \underline{59.33}
& \underline{60.22}
& \underline{55.22}
& \underline{56.11}
& \textbf{40.89}
& \textbf{67.22}
& \underline{56.50}
& 8.74
& \ums{27.09}{4.76}
& \bms{24.35}{4.90} \\

GPT-4
& 18.00
& 22.73
& 27.82
& 15.45
& 20.46
& 39.09
& 23.93
& 8.55
& \bms{4.42}{1.20}
& \ums{27.35}{3.49} \\

StyleTunedLM
& 56.56
& 58.11
& 52.11
& 55.11
& 35.56
& 62.89
& 53.39
& 9.43
& \ms{40.38}{5.42}
& \ms{37.48}{5.60} \\
\bottomrule
\end{tabular}
\end{table}

\paragraph{Qualitative analysis.}
Fig.~\ref{fig:appendix_prefix_grid_london} presents an example from the general SI setting. The target author is Jack London, while the prefix source is Herbert George Wells, creating a cross-genre condition between adventure/survival writing and speculative science fiction. %
The generation adapts London-like style and themes to a less stereotypical content setting. The generated text is classified as Jack London by all four style evaluators, but its overlap with Jack London's training passages remains low according to longest copied span. This suggests that \EDITORS w/o $\Delta W^{(0)}$ can use an externally supplied prefix to control the generation start while still expressing the target author's style, rather than relying on direct memorization of target-author training text.

\begin{figure*}[t]
\centering
\fbox{
\begin{minipage}{0.96\textwidth}
\small
\textbf{Generation sample: prefix-continuation SI.}

\vspace{0.4em}
\begin{tabular}{ll}
\textbf{Target author} & Jack London \\
\textbf{Prefix source author} & Herbert George Wells \\
\textbf{Prefix} & ``The third time I got'' \\
\textbf{Style similarity} & 4/4 style evaluators predict the target author \\
\textbf{Longest copied span} & 6 words \\
\end{tabular}

\vspace{0.8em}
\textbf{Generated passage excerpt.}
\begin{quote}
\small
\emph{
``The third time I got 5,000 words written. It was a good piece of writing, and I was proud of it. But it was my last piece. I was tired. I was worn out from the strain of the last three weeks. I had been up at 4:30 A.M. every day, and, with the exception of two or three days when I had been unable to work, I had kept at it until 8 P.M. Sometimes I had been able to keep at it until 10 P.M. or later. \ldots{} I had made my fight, and I had won. I had beaten the devil, and had gained the day. And I was tired. \ldots{} I had won the fight, and I was tired.''}
\end{quote}

\vspace{0.2em}
\end{minipage}
}
\caption{Generation sample for general SI.}
\label{fig:appendix_prefix_grid_london}
\end{figure*}

\subsection{Content-controlled SI Results}
\label{app:main_result}

\cref{tab:style_similarity_all_evaluators}, \ref{tab:independent_style_evaluation}, \ref{tab:content_evaluation_robustness}, \ref{tab:memorization_audit} and \ref{tab:style_imitation_results} give a more comprehensive version of the content-controlled SI results shown in \cref{fig:si_style} and \cref{fig:si_results_table} in the main paper. 
We report averages over three independent generation trials using random seeds 1001, 1002, and 1003. The same generated outputs from each trial are evaluated by all automatic metrics. For evaluator-specific style similarity, each evaluator column reports the mean target-author recognition rate across the three trials, with the value after $\pm$ denoting the trial standard deviation. The reported Style Similarity average is computed by first averaging the six evaluator means, and its trial standard deviation is propagated as $\sqrt{\sum_j \sigma_j^2}/6$. The quantity $\sigma_{\mathrm{eval}}$ measures disagreement among the six evaluator mean scores and should not be interpreted as trial uncertainty. For the aggregate SI table, all reported standard deviations are computed across the same three generation trials. 

\begin{table}[H]
\centering
\caption{Style similarity performance on Gutenberg30 across six evaluators. Each evaluator column reports target-author recognition rate (\%). Avg. is the mean across the six evaluator means. $\sigma_{\mathrm{eval}}$ reflects evaluator disagreement. \textbf{Bold} and \underline{underline} mark the best and second-best values, respectively.}
\small
\setlength{\tabcolsep}{1.5pt}
\renewcommand{\arraystretch}{0.78}
\newcommand{\mstd}[2]{\ensuremath{\text{#1}{\scriptsize\pm}{\scriptsize \text{#2}}}}
\newcommand{\ms}[2]{#1{\scriptsize$\pm$#2}}
\newcommand{\bms}[2]{\textbf{#1}{\scriptsize$\pm$#2}}
\newcommand{\ums}[2]{\underline{#1}{\scriptsize$\pm$#2}}
\begin{tabular}{lcccccccc}
\toprule
\textbf{Model} 
& \textbf{Llama} 
& \textbf{Gemma} 
& \textbf{Phi4} 
& \textbf{Mistral} 
& \textbf{STAR} 
& \textbf{BertAA} 
& \textbf{Avg.} 
& $\boldsymbol{\sigma_{\mathrm{eval}}}$ \\
\midrule

GPT-4 
& \mstd{8.73}{1.12} 
& \mstd{5.56}{1.37} 
& \mstd{18.41}{0.59} 
& \mstd{5.40}{0.98} 
& \mstd{9.21}{0.59} 
& \mstd{17.62}{1.78} 
& \mstd{10.82}{0.47} 
& 5.79 \\

ASTRAPOP 
& \mstd{52.54}{2.00} 
& \mstd{53.49}{0.59} 
& \mstd{52.38}{0.00} 
& \mstd{52.06}{1.84} 
& \mstd{25.08}{1.37} 
& \mstd{40.63}{1.25} 
& \mstd{46.03}{0.56} 
& 11.34 \\

StoryTrans 
& \mstd{8.89}{0.98} 
& \mstd{5.87}{0.59} 
& \mstd{15.08}{0.59} 
& \mstd{10.95}{2.16} 
& \mstd{13.97}{1.47} 
& \mstd{25.87}{1.19} 
& \mstd{13.44}{0.52} 
& 6.95 \\

StyleTunedLM 
& \mstd{54.13}{2.97} 
& \mstd{50.00}{2.72} 
& \mstd{46.35}{1.25} 
& \mstd{52.86}{0.67} 
& \mstd{22.06}{2.59} 
& \mstd{40.63}{2.97} 
& \mstd{44.34}{0.97} 
& 11.96 \\
\midrule

EDITORS w/o $\Delta W^{(0)}$ 
& \bms{56.67}{0.67}
& \bms{63.33}{1.40} 
& \ums{55.08}{2.34} 
& \bms{56.83}{1.12} 
& \bms{36.19}{0.45} 
& \bms{62.38}{1.35} 
& \bms{55.08}{0.56}
& 9.84 \\

EDITORS w/ $\Delta W^{(0)}$ 
& \ums{56.19}{2.43} 
& \ums{60.16}{2.65} 
& \bms{56.67}{1.03} 
& \bms{56.83}{1.12} 
& \ums{33.81}{0.53} 
& \ums{60.95}{0.98} 
& \ums{54.10}{0.68} 
& 10.14 \\
\bottomrule
\end{tabular}
\label{tab:style_similarity_all_evaluators}
\end{table}

\begin{table}[H]
\centering
\caption{Supplementary style similarity performance on Gutenberg30 using independent third-party evaluators StyleDistance and StyleEmbedding. \textbf{Bold} and \underline{underline} mark the best and second-best values, respectively.}
\label{tab:independent_style_evaluation}
\small
\providecommand{\ms}[2]{#1{\scriptsize$\pm$#2}}
\providecommand{\bms}[2]{\textbf{#1}{\scriptsize$\pm$#2}}
\providecommand{\ums}[2]{\underline{#1}{\scriptsize$\pm$#2}}
\begin{tabular}{lccc}
\toprule
\textbf{Model} & \textbf{StyleDistance} & \textbf{StyleEmbedding} \\
\midrule
GPT-4
& \ms{4.92}{1.10}
& \ms{6.03}{0.55} \\
StoryTrans
& \ms{9.37}{0.99}
& \ms{7.94}{1.45} \\
ASTRAPOP
& \ums{17.30}{1.45}
& \ums{14.60}{2.44} \\
StyleTunedLM
& \ms{7.78}{2.15}
& \ms{11.11}{1.10} \\
EDITORS
& \bms{22.70}{2.15}
& \bms{19.52}{2.65} \\
\bottomrule
\end{tabular}
\end{table}

\begin{table}[H]
\centering
\caption{Content-preservation evaluation on Gutenberg30 using
two LLM judges and NLI-based entailment rate~\citep{chen-eger-2023-menli} as a classical yet less powerful alternative. \textbf{Bold} and \underline{underline} indicate the best and
second-best means, respectively.}
\label{tab:content_evaluation_robustness}
\small

\providecommand{\ms}[2]{#1{\scriptsize$\pm$#2}}
\providecommand{\bms}[2]{\textbf{#1}{\scriptsize$\pm$#2}}
\providecommand{\ums}[2]{\underline{#1}{\scriptsize$\pm$#2}}

\begin{tabular}{lccc}
\toprule
\textbf{Model} & \textbf{GPT-4o mini} & \textbf{Qwen3.5-27B} & \textbf{NLI-based entailment rate} \\
\midrule
GPT-4
& \bms{100.00}{0.00}
& \bms{99.76}{0.12}
& \bms{16.18}{3.23} \\
StoryTrans
& \ms{0.83}{1.34}
& \ms{0.52}{0.89}
& \ms{3.49}{2.79} \\
ASTRAPOP
& \ms{4.40}{0.48}
& \ms{3.06}{0.50}
& \ms{3.65}{1.10} \\
StyleTunedLM
& \ms{66.79}{1.49}
& \ms{52.46}{1.19}
& \ms{4.92}{1.67} \\
EDITORS
& \ums{72.54}{1.35}
& \ums{63.77}{1.19}
& \ums{10.48}{0.48} \\
\bottomrule
\end{tabular}
\end{table}

\begin{table}[H]
\centering
\caption{Character-level training-corpus overlap on Gutenberg30. \textbf{Bold} and \underline{underline} indicate the best and second-best
results, respectively.}
\label{tab:memorization_audit}
\small
\setlength{\tabcolsep}{4pt}

\providecommand{\ms}[2]{#1{\scriptsize$\pm$#2}}
\providecommand{\bms}[2]{\textbf{#1}{\scriptsize$\pm$#2}}
\providecommand{\ums}[2]{\underline{#1}{\scriptsize$\pm$#2}}

\begin{tabular}{lcccc}
\toprule
\textbf{Model}
& \textbf{\shortstack{OR@25 (\%)\\Target author}}
& \textbf{\shortstack{OR@25 (\%)\\Global corpus}}
& \textbf{\shortstack{Maximum copied span \\ (characters)}} \\
\midrule
GPT-4
& \ums{0.10}{0.12}
& \ums{0.94}{0.04}
& \underline{39} \\
StoryTrans
& \bms{0.01}{0.01}
& \bms{0.09}{0.04}
& \textbf{25} \\
ASTRAPOP
& \ms{1.74}{0.39}
& \ms{6.88}{0.32}
& 1934 \\
StyleTunedLM
& \ms{1.11}{0.04}
& \ms{6.32}{0.33}
& 51 \\
EDITORS
& \ms{0.74}{0.01}
& \ms{4.64}{0.25}
& 44 \\
\bottomrule
\end{tabular}
\end{table}

\begin{table}[H]
\centering
\caption{Content-controlled SI performance on Gutenberg30. Style Similarity and Content Preservation are higher-is-better metrics. BLEU and Self-BLEU are reported on a 0--100 scale, and LCS$_{\max}$ is measured in words; lower is better for these memorization/diversity diagnostics. \textbf{Bold} and \underline{underline} mark the best and second-best values, respectively.}
\label{tab:style_imitation_results}
\small
\setlength{\tabcolsep}{3.0pt}
\renewcommand{\arraystretch}{0.8}
\newcommand{\ms}[2]{#1{\scriptsize$\pm$#2}}
\newcommand{\bms}[2]{\textbf{#1}{\scriptsize$\pm$#2}}
\newcommand{\ums}[2]{\underline{#1}{\scriptsize$\pm$#2}}
\begin{tabular}{lccccc}
\toprule
\textbf{Model} 
& \textbf{Style Sim. (\%)} 
& \textbf{Content Pres.} 
& \textbf{BLEU} 
& \textbf{Self-BLEU} 
& \textbf{LCS$_{\max}$} \\
\midrule
GPT-4 
& \ms{10.82}{0.47} 
& \bms{99.88}{0.06} 
& \ums{16.54}{0.12} 
& \bms{11.03}{0.13} 
& \ums{9.67}{1.53} \\

ASTRAPOP 
& \ms{46.03}{0.56} 
& \ms{3.73}{0.49} 
& \ms{27.08}{0.30}
& \ms{14.85}{0.50} 
& \ms{161.00}{240.92} \\

StoryTrans 
& \ms{13.44}{0.52} 
& \ms{0.68}{1.12} 
& \bms{4.96}{0.95} 
& \ms{14.36}{0.38} 
& \bms{9.00}{0.58} \\

StyleTunedLM 
& \ms{44.34}{0.97} 
& \ms{59.63}{1.34} 
& \ms{47.18}{0.16}
& \ms{20.54}{0.39} 
& \ms{12.00}{0.00} \\

\midrule
EDITORS w/ $\Delta W^{(0)}$ 
& \ums{54.10}{0.68} 
& \ms{39.23}{2.54} 
& \ms{45.64}{0.26} 
& \ms{21.07}{0.43} 
& \ms{14.33}{1.53} \\

EDITORS w/o $\Delta W^{(0)}$ 
& \bms{55.08}{0.56} 
& \ums{68.16}{1.27} 
& \ms{27.42}{0.64} 
& \ums{13.14}{0.55} 
& \ms{12.67}{1.53} \\
\bottomrule
\end{tabular}
\end{table}

\paragraph{Qualitative analysis.}
Fig.~\ref{fig:si_sample_garland} illustrates a generation sample from \EDITORS w/o $\Delta W^{(0)}$. It shows that the model can preserve an adventure/survival content condition while producing text that is recognized as Hamlin Garland by all style evaluators. The generated passage  transfers Garland-like social observation, first-person plainness into a survival setting. 

\begin{figure}[H]
\centering
\fbox{
\begin{minipage}{0.96\textwidth}
\small
\textbf{Generation sample: content-controlled SI.}

\vspace{0.4em}
\begin{tabular}{ll}
\textbf{Target author} & Hamlin Garland \\
\textbf{Author's dominant genre} & Social realism / local color (D) \\
\textbf{Generated content pack} & A group works together to survive after getting stranded. \\
\textbf{Generated content genre} & Adventure / survival / exploration (B) \\
\textbf{Content preservation} & 5/5, label: yes \\
\textbf{Style similarity} & 6/6, all style evaluators predict the target author \\
\end{tabular}

\vspace{0.8em}
\textbf{Generated passage excerpt.}
\begin{quote}
\small
\emph{
``We were all in this venture together, and every man was willing to risk his life for the sake of the others, and the only question was as to the division of the spoils. In this matter the two Canadians were very shrewd and took a leading part in the discussion, while the three of us who were Americans were content to take the back seats and let them do the talking. \ldots{} We were a band of outcasts, but we were outcasts who had made a new social compact. We were a band of outcasts who had come together for the purpose of mutual protection. At this time I was a tramp, and I had no home to go to, but I was a tramp who had a home.''}
\end{quote}

\vspace{0.2em}
\end{minipage}
}
\caption{Generation sample under a cross-genre content specification.}
\label{fig:si_sample_garland}
\end{figure}

\paragraph{SI on Blogs10.}
To evaluate SI beyond literary texts, we compare \EDITORS with StyleTunedLM (baseline with best performance) on Blogs10. As shown in \cref{tab:si_blogs10}, \EDITORS achieves higher style similarity and content preservation, together with lower training-corpus overlap measured by BLEU-4 and $\mathrm{OR}@25$, and greater output diversity measured by lower Self-BLEU. These results suggest that \EDITORS' advantage does not depend on styles being strongly distinctive, the benefits of \EDITORS also extend to blog posts.

\begin{table}[H]
\centering
\small
\setlength{\tabcolsep}{4pt}
\providecommand{\ms}[2]{#1{\scriptsize$\pm$#2}}
\providecommand{\bms}[2]{\textbf{#1}{\scriptsize$\pm$#2}}
\caption{Content-controlled SI performance on Blogs10.
Arrows indicate the preferred direction.
\textbf{Bold} marks the better mean for each metric.}
\label{tab:si_blogs10}
\begin{tabular}{lccccc}
\toprule
\textbf{Method}
& \shortstack{\textbf{Style Sim. (\%)} $\uparrow$}
& \shortstack{\textbf{Content Pres. } $\uparrow$}
& \textbf{BLEU-4} $\downarrow$
& \textbf{Self-BLEU} $\downarrow$
& \textbf{OR@25} $\downarrow$ \\
\midrule
StyleTunedLM
& \ms{22.14}{1.00}
& \ms{86.43}{1.30}
& \ms{36.62}{1.60}
& \ms{15.50}{0.52}
& \ms{0.92}{0.14} \\
\EDITORS
& \bms{37.62}{1.33}
& \bms{94.29}{0.36}
& \bms{35.02}{1.52}
& \bms{11.74}{1.38}
& \bms{0.75}{0.20} \\
\bottomrule
\end{tabular}
\end{table}
\subsection{Content Retention}
\label{app:content_probe}

To construct the QA dataset, we prompt GPT-4o mini to generate one factual question-answer pair from each original passage. The question must be answerable by copying a single explicit statement from the facts, avoiding yes/no questions. We then prompt each model variant ($W$, $\Delta W^{(c)}$, $\Delta W^{(0)}$, $\Delta W^{(s)}$) to answer the question and evaluate correctness with both ROUGE and an LLM judge.

\paragraph{QA pair construction.}
We prompt GPT-4o mini with the style-neutral statements from each passage to generate one factual question-answer pair per passage.

\begin{tcolorbox}[colback=gray!5, colframe=gray!50]
\small
\texttt{You are generating a dataset of question-answer pairs.}\\
\texttt{Create exactly one question that can be answered by copying a single explicit statement from the passage.}\\
\texttt{Avoid yes/no questions.}\\
\texttt{Make it specific and factual.}\\
\texttt{Output JSON only with keys "question" and "answer".}\\[4pt]
\texttt{style-neutral statements: \{text\}}
\end{tcolorbox}

\paragraph{LLM-as-judge evaluation.}

For each model-generated answer, we use GPT-4o mini to judge semantic correctness against the gold answer. 

\begin{tcolorbox}[colback=gray!5, colframe=gray!50]
\small
\texttt{Judge whether the model answer means the same thing as the gold answer.}\\
\texttt{Be strict: mark correct only if the meaning matches.}\\
\texttt{Return JSON only with keys: label, score.}\\
\texttt{label must be one of: correct, incorrect.}\\
\texttt{score must be 1 for correct and 0 for incorrect.}\\[4pt]
\texttt{Question: \{question\}}\\
\texttt{Gold answer: \{gold\}}\\
\texttt{Model answer: \{pred\}}
\end{tcolorbox}

\paragraph{Results on different base models.}
\begin{figure}[H]
\centering
\vspace{-1em}
\begin{tikzpicture}
\begin{groupplot}[
    group style={
        group size=4 by 1,
        horizontal sep=0.6cm,
    },
    ybar,
    width=0.28\textwidth,
    height=4cm,
    ymin=0, ymax=100,
    symbolic x coords={$W$,$\Delta W^{(c)}$,$\Delta W^{(0)}$,$\Delta W^{(s)}$},
    xtick=data,
    enlarge x limits=0.18,
    grid=major,
    grid style={dashed,gray!25},
    x tick label style={font=\scriptsize, rotate=25, anchor=east},
    y tick label style={font=\scriptsize},
    title style={font=\small},
    ylabel style={font=\small},
    every axis plot/.append style={bar width=4pt},
]
\nextgroupplot[title={Llama-3.1-8B}, ylabel={Score (\%)}]
\addplot+[bar shift=-3pt, fill=blue!60, draw=blue!80] coordinates {
  ($W$,58.84) ($\Delta W^{(c)}$,78.39) ($\Delta W^{(0)}$,60.11) ($\Delta W^{(s)}$,54.54)};
\addplot+[bar shift=3pt, fill=orange!60, draw=orange!80] coordinates {
  ($W$,63.22) ($\Delta W^{(c)}$,82.61) ($\Delta W^{(0)}$,67.72) ($\Delta W^{(s)}$,61.77)};

\nextgroupplot[title={Gemma2-9B}, yticklabels={}]
\addplot+[bar shift=-3pt, fill=blue!60, draw=blue!80] coordinates {
  ($W$,64.56) ($\Delta W^{(c)}$,83.09) ($\Delta W^{(0)}$,65.76) ($\Delta W^{(s)}$,61.42)};
\addplot+[bar shift=3pt, fill=orange!60, draw=orange!80] coordinates {
  ($W$,64.67) ($\Delta W^{(c)}$,84.67) ($\Delta W^{(0)}$,69.00) ($\Delta W^{(s)}$,62.33)};

\nextgroupplot[title={Mistral-7B}, yticklabels={}]
\addplot+[bar shift=-3pt, fill=blue!60, draw=blue!80] coordinates {
  ($W$,55.03) ($\Delta W^{(c)}$,69.26) ($\Delta W^{(0)}$,53.46) ($\Delta W^{(s)}$,51.18)};
\addplot+[bar shift=3pt, fill=orange!60, draw=orange!80] coordinates {
  ($W$,51.18) ($\Delta W^{(c)}$,65.00) ($\Delta W^{(0)}$,49.33) ($\Delta W^{(s)}$,47.33)};

\nextgroupplot[title={Phi4-mini-3.8B}, yticklabels={}]
\addplot+[bar shift=-3pt, fill=blue!60, draw=blue!80] coordinates {
  ($W$,48.16) ($\Delta W^{(c)}$,61.10) ($\Delta W^{(0)}$,46.60) ($\Delta W^{(s)}$,29.20)};
\addplot+[bar shift=3pt, fill=orange!60, draw=orange!80] coordinates {
  ($W$,43.33) ($\Delta W^{(c)}$,59.33) ($\Delta W^{(0)}$,44.67) ($\Delta W^{(s)}$,23.67)};

\end{groupplot}

\node[anchor=north, yshift=-0.8cm] at ($(group c2r1.south)!0.5!(group c3r1.south)$) {
  \begin{tikzpicture}
    \draw[fill=blue!60, draw=blue!80] (0,0) rectangle (0.25,0.15);
    \node[right, font=\scriptsize] at (0.3,0.075) {QA ROUGE};
    \draw[fill=orange!60, draw=orange!80] (1.8,0) rectangle (2.05,0.15);
    \node[right, font=\scriptsize] at (2.1,0.075) {LLM-as-Judge};
  \end{tikzpicture}
};

\end{tikzpicture}
\vspace{-1em}
\caption{Content retention across four backbone LLMs. }
\label{fig:content_probe}
\vspace{-1em}
\end{figure}

Figure~\ref{fig:content_probe} shows two consistent patterns across all four backbones. First, $\Delta W^{(c)}$ always scores highest, confirming that the SVD in \cref{sec:method_diagnosis} captures the dominant semantic signal. Second, $\Delta W^{(s)}$ consistently scores below $\Delta W^{(0)}$, indicating that activation-space deflation reduces reliance on content-oriented directions slightly in the style adapter while still retaining sufficient residual semantic information for attribution. %

\subsection{Style Retention in $\mathcal{D}^{(c)}$}
To assess the remaining author-identifying information, we evaluate AA accuracy on the style-neutral statements $\mathcal{D}^{(c)}$ and the original passages $\mathcal{D}$ using \EDITORS based on Llama-3.1-8B-Instruct.
As shown in \cref{tab:aa_extracted_facts}, accuracy decreases
substantially across all three datasets. These results suggest
that style-neutral statements extraction reduces author-identifying information, although residual style and content cues can remain.

\begin{table}[H]
\centering
\small
\caption{AA accuracy (\%) on extracted facts and original
passages, evaluated using \EDITORS based on
Llama-3.1-8B-Instruct.}
\label{tab:aa_extracted_facts}
\begin{tabular}{lcc}
\toprule
\textbf{Dataset}
& \textbf{AA accuracy on $\mathcal{D}^{(c)}$}
& \textbf{AA accuracy on $\mathcal{D}$} \\
\midrule
CCAT50      & 52.80 & 80.40 \\
Blogs50     & 33.67 & 83.80 \\
Gutenberg30 & 27.17 & 82.33 \\
\bottomrule
\end{tabular}
\end{table}

\subsection{Activation- vs.\ Parameter-Space Deflation}
\label{app:posthoc}

\paragraph{Post-hoc deflation.}
Prior work on deflation in LoRA typically applies projection to the
adapter weights rather than the activations~\citep{chen2026nploranullspaceprojection}. We test whether a similar post-hoc parameter-space projection
can replace our activation-space deflation. Concretely, after training a
domain warm-up adapter $\Delta W^{(0)}_{\ell}$ for the same total budget ($8$ epochs) as
EDITORS, we right-project it onto the null space of the diagnosed content basis:
\begin{equation*}
\Delta W^{(s)}_{\ell} \;\leftarrow\; \Delta W^{(0)}_{\ell}\, P_{\ell}^{\perp},
\quad\text{where}\quad
P_{\ell}^{\perp} = I - V_{\ell}V_{\ell}^\top.
\label{eq:posthoc}
\end{equation*}
No further fine-tuning is applied; the projected adapter is used directly for
attribution. This baseline isolates the effect of \emph{when} the content
subspace is removed: during optimization (\EDITORS) versus after a
content-entangled adapter has already been trained.

\paragraph{Parameter-space orthogonal LoRA.}
Prior orthogonal LoRA and null-space editing methods impose constraints directly on parameter updates to reduce interference across tasks or preserve existing knowledge~\citep{wang2023orthogonal,fang2025alphaedit}. Motivated by this family of methods, we implement the training-time parameter-space orthogonal projection that constrains the final style LoRA update to be orthogonal to the content subspace. For each target module $(\ell,\tau)$, we construct the protected basis $V_{\ell,\tau}^{(c)}$ from the content adapter and project the trainable style LoRA $A$-matrix before training and after every optimizer step:
\[
A_{\ell,\tau}^{(s)}
\leftarrow
A_{\ell,\tau}^{(s)}
-
\left(A_{\ell,\tau}^{(s)}V_{\ell,\tau}^{(c)}\right)
\left(V_{\ell,\tau}^{(c)}\right)^\top .
\]
This enforces
\[
A_{\ell,\tau}^{(s)}V_{\ell,\tau}^{(c)}=0,
\qquad
\Delta W_{\ell,\tau}^{(s)}V_{\ell,\tau}^{(c)}
=
B_{\ell,\tau}^{(s)}A_{\ell,\tau}^{(s)}V_{\ell,\tau}^{(c)}
=
0,
\]
so the newly trained style update cannot act along the protected
content-oriented input directions.

\begin{figure}[H]
\centering
\begin{tikzpicture}
\begin{axis}[
    ybar,
    bar width=8pt,
    width=0.92\linewidth,
    height=5.4cm,
    ymin=0, ymax=100,
    ylabel={Accuracy (\%)},
    symbolic x coords={CCAT50,Blogs50,Gutenberg30},
    xtick=data,
    enlarge x limits=0.24,
    grid=major,
    grid style={dashed,gray!30},
    legend style={
        at={(0.5,1.12)},
        anchor=south,
        legend columns=2,
        draw=none,
        font=\small,
        /tikz/every even column/.append style={column sep=8pt}
    },
    nodes near coords,
    nodes near coords align={vertical},
    every node near coord/.append style={font=\tiny, yshift=1pt},
]
\addplot+[bar shift=-15pt] coordinates {
    (CCAT50,80.40)
    (Blogs50,83.80)
    (Gutenberg30,82.33)
};
\addplot+[bar shift=-5pt] coordinates {
    (CCAT50,73.00)
    (Blogs50,75.49)
    (Gutenberg30,76.33)
};
\addplot+[bar shift=5pt] coordinates {
    (CCAT50,67.33)
    (Blogs50,65.33)
    (Gutenberg30,75.47)
};
\addplot+[bar shift=15pt] coordinates {
    (CCAT50,51.00)
    (Blogs50,28.33)
    (Gutenberg30,53.00)
};
\legend{EDITORS (activation-space), Parameter-space orthogonal LoRA, Plain SFT, Post-hoc projection}
\end{axis}
\end{tikzpicture}

\caption{Comparison of activation-space and parameter-space alternatives on three AA benchmarks. \EDITORS consistently outperforms plain SFT, the post-hoc projection baseline, and the parameter-space orthogonal LoRA baseline.}
\label{fig:projection_ablation}
\vspace{-1em}
\end{figure}

\paragraph{Activation-space vs. parameter-space deflation.}
\cref{fig:projection_ablation} compares \EDITORS with two parameter-space alternatives: a \emph{post-hoc projection} baseline and a \emph{parameter-space orthogonal LoRA} baseline. The post-hoc projection baseline degrades accuracy substantially on all three datasets relative to \EDITORS, with the largest gap on Blogs50. This suggests that once an adapter has already absorbed a mixture of style, content, and domain priors, projecting its weights after training removes not only content shortcuts but also author-discriminative signals entangled with those directions. In contrast, \EDITORS applies deflation during optimization, so the style adapter learns from the beginning under restricted access to content-dominated directions rather than being destructively rewritten after training.

The parameter-space orthogonal LoRA baseline also underperforms \EDITORS and yields smaller gains over plain SFT. This indicates that constraining only the new LoRA update is not sufficient to disentangle style from content. Although the orthogonality constraint enforces
$\Delta W_{\ell,\tau}^{(s)} V_{\ell,\tau}^{(c)} = 0$,
the frozen base path still receives the full activation \(x_\ell\), so semantic information can still leak through the backbone. In other words, the constraint is too soft: it limits where the new LoRA update can write, but it does not directly control what information is available during the forward pass. By contrast, \EDITORS removes the protected component from the hidden activation itself,
\[
x_{\ell,\tau}^{\mathrm{def}}
=
\left(I-V_{\ell,\tau}^{(c)}(V_{\ell,\tau}^{(c)})^\top\right)x_\ell,
\]
so both the frozen base computation and the style LoRA branch are prevented from using the protected content directions at the selected module. This makes activation-space deflation a stronger and more direct intervention. It also aligns better with our modeling assumption that hidden representations contain partially separable content-oriented and style-oriented directions. 

Finally, activation-space deflation is not limited to LoRA parameterization: the same intervention can in principle be applied to other adaptation methods, making it the more flexible design choice overall.

\newpage
\section*{NeurIPS Paper Checklist}

\begin{enumerate}

\item {\bf Claims}
    \item[] Question: Do the main claims made in the abstract and introduction accurately reflect the paper's contributions and scope?
    \item[] Answer: \answerYes{} %
    \item[] Justification: In \cref{sec:introduction}, we include the corresponding section number for each claim we make.
    \item[] Guidelines:
    \begin{itemize}
        \item The answer \answerNA{} means that the abstract and introduction do not include the claims made in the paper.
        \item The abstract and/or introduction should clearly state the claims made, including the contributions made in the paper and important assumptions and limitations. A \answerNo{} or \answerNA{} answer to this question will not be perceived well by the reviewers. 
        \item The claims made should match theoretical and experimental results, and reflect how much the results can be expected to generalize to other settings. 
        \item It is fine to include aspirational goals as motivation as long as it is clear that these goals are not attained by the paper. 
    \end{itemize}

\item {\bf Limitations}
    \item[] Question: Does the paper discuss the limitations of the work performed by the authors?
    \item[] Answer: \answerYes{} %
    \item[] Justification: Refer to Sec.~\ref{sec:conclusion}.
    \item[] Guidelines:
    \begin{itemize}
        \item The answer \answerNA{} means that the paper has no limitation while the answer \answerNo{} means that the paper has limitations, but those are not discussed in the paper. 
        \item The authors are encouraged to create a separate ``Limitations'' section in their paper.
        \item The paper should point out any strong assumptions and how robust the results are to violations of these assumptions (e.g., independence assumptions, noiseless settings, model well-specification, asymptotic approximations only holding locally). The authors should reflect on how these assumptions might be violated in practice and what the implications would be.
        \item The authors should reflect on the scope of the claims made, e.g., if the approach was only tested on a few datasets or with a few runs. In general, empirical results often depend on implicit assumptions, which should be articulated.
        \item The authors should reflect on the factors that influence the performance of the approach. For example, a facial recognition algorithm may perform poorly when image resolution is low or images are taken in low lighting. Or a speech-to-text system might not be used reliably to provide closed captions for online lectures because it fails to handle technical jargon.
        \item The authors should discuss the computational efficiency of the proposed algorithms and how they scale with dataset size.
        \item If applicable, the authors should discuss possible limitations of their approach to address problems of privacy and fairness.
        \item While the authors might fear that complete honesty about limitations might be used by reviewers as grounds for rejection, a worse outcome might be that reviewers discover limitations that aren't acknowledged in the paper. The authors should use their best judgment and recognize that individual actions in favor of transparency play an important role in developing norms that preserve the integrity of the community. Reviewers will be specifically instructed to not penalize honesty concerning limitations.
    \end{itemize}

\item {\bf Theory assumptions and proofs}
    \item[] Question: For each theoretical result, does the paper provide the full set of assumptions and a complete (and correct) proof?
    \item[] Answer: \answerNA{} %
    \item[] Justification: \answerNA{}
    \item[] Guidelines:
    \begin{itemize}
        \item The answer \answerNA{} means that the paper does not include theoretical results. 
        \item All the theorems, formulas, and proofs in the paper should be numbered and cross-referenced.
        \item All assumptions should be clearly stated or referenced in the statement of any theorems.
        \item The proofs can either appear in the main paper or the supplemental material, but if they appear in the supplemental material, the authors are encouraged to provide a short proof sketch to provide intuition. 
        \item Inversely, any informal proof provided in the core of the paper should be complemented by formal proofs provided in appendix or supplemental material.
        \item Theorems and Lemmas that the proof relies upon should be properly referenced. 
    \end{itemize}

    \item {\bf Experimental result reproducibility}
    \item[] Question: Does the paper fully disclose all the information needed to reproduce the main experimental results of the paper to the extent that it affects the main claims and/or conclusions of the paper (regardless of whether the code and data are provided or not)?
    \item[] Answer: \answerYes{} %
    \item[] Justification: Refer to App.~\ref{app:implementation} and~\ref{app:prompts} and the supplemental material submitted.
    \item[] Guidelines:
    \begin{itemize}
        \item The answer \answerNA{} means that the paper does not include experiments.
        \item If the paper includes experiments, a \answerNo{} answer to this question will not be perceived well by the reviewers: Making the paper reproducible is important, regardless of whether the code and data are provided or not.
        \item If the contribution is a dataset and\slash or model, the authors should describe the steps taken to make their results reproducible or verifiable. 
        \item Depending on the contribution, reproducibility can be accomplished in various ways. For example, if the contribution is a novel architecture, describing the architecture fully might suffice, or if the contribution is a specific model and empirical evaluation, it may be necessary to either make it possible for others to replicate the model with the same dataset, or provide access to the model. In general. releasing code and data is often one good way to accomplish this, but reproducibility can also be provided via detailed instructions for how to replicate the results, access to a hosted model (e.g., in the case of a large language model), releasing of a model checkpoint, or other means that are appropriate to the research performed.
        \item While NeurIPS does not require releasing code, the conference does require all submissions to provide some reasonable avenue for reproducibility, which may depend on the nature of the contribution. For example
        \begin{enumerate}
            \item If the contribution is primarily a new algorithm, the paper should make it clear how to reproduce that algorithm.
            \item If the contribution is primarily a new model architecture, the paper should describe the architecture clearly and fully.
            \item If the contribution is a new model (e.g., a large language model), then there should either be a way to access this model for reproducing the results or a way to reproduce the model (e.g., with an open-source dataset or instructions for how to construct the dataset).
            \item We recognize that reproducibility may be tricky in some cases, in which case authors are welcome to describe the particular way they provide for reproducibility. In the case of closed-source models, it may be that access to the model is limited in some way (e.g., to registered users), but it should be possible for other researchers to have some path to reproducing or verifying the results.
        \end{enumerate}
    \end{itemize}

\item {\bf Open access to data and code}
    \item[] Question: Does the paper provide open access to the data and code, with sufficient instructions to faithfully reproduce the main experimental results, as described in supplemental material?
    \item[] Answer: \answerYes{} %
    \item[] Justification: Refer to the submitted code and datasets.
    \item[] Guidelines:
    \begin{itemize}
        \item The answer \answerNA{} means that paper does not include experiments requiring code.
        \item Please see the NeurIPS code and data submission guidelines (\url{https://neurips.cc/public/guides/CodeSubmissionPolicy}) for more details.
        \item While we encourage the release of code and data, we understand that this might not be possible, so \answerNo{} is an acceptable answer. Papers cannot be rejected simply for not including code, unless this is central to the contribution (e.g., for a new open-source benchmark).
        \item The instructions should contain the exact command and environment needed to run to reproduce the results. See the NeurIPS code and data submission guidelines (\url{https://neurips.cc/public/guides/CodeSubmissionPolicy}) for more details.
        \item The authors should provide instructions on data access and preparation, including how to access the raw data, preprocessed data, intermediate data, and generated data, etc.
        \item The authors should provide scripts to reproduce all experimental results for the new proposed method and baselines. If only a subset of experiments are reproducible, they should state which ones are omitted from the script and why.
        \item At submission time, to preserve anonymity, the authors should release anonymized versions (if applicable).
        \item Providing as much information as possible in supplemental material (appended to the paper) is recommended, but including URLs to data and code is permitted.
    \end{itemize}

\item {\bf Experimental setting/details}
    \item[] Question: Does the paper specify all the training and test details (e.g., data splits, hyperparameters, how they were chosen, type of optimizer) necessary to understand the results?
    \item[] Answer: \answerYes{} %
    \item[] Justification: Refer to App.~\ref{app:dataset_stats} and~\ref{app:implementation}.
    \item[] Guidelines:
    \begin{itemize}
        \item The answer \answerNA{} means that the paper does not include experiments.
        \item The experimental setting should be presented in the core of the paper to a level of detail that is necessary to appreciate the results and make sense of them.
        \item The full details can be provided either with the code, in appendix, or as supplemental material.
    \end{itemize}

\item {\bf Experiment statistical significance}
    \item[] Question: Does the paper report error bars suitably and correctly defined or other appropriate information about the statistical significance of the experiments?
    \item[] Answer: \answerYes{} %
    \item[] Justification: Refer to App.~\ref{app:ablations},~\ref{app:aa_implementation} and Sec.~\ref{sec:style_imitation}.
    \item[] Guidelines:
    \begin{itemize}
        \item The answer \answerNA{} means that the paper does not include experiments.
        \item The authors should answer \answerYes{} if the results are accompanied by error bars, confidence intervals, or statistical significance tests, at least for the experiments that support the main claims of the paper.
        \item The factors of variability that the error bars are capturing should be clearly stated (for example, train/test split, initialization, random drawing of some parameter, or overall run with given experimental conditions).
        \item The method for calculating the error bars should be explained (closed form formula, call to a library function, bootstrap, etc.)
        \item The assumptions made should be given (e.g., Normally distributed errors).
        \item It should be clear whether the error bar is the standard deviation or the standard error of the mean.
        \item It is OK to report 1-sigma error bars, but one should state it. The authors should preferably report a 2-sigma error bar than state that they have a 96\% CI, if the hypothesis of Normality of errors is not verified.
        \item For asymmetric distributions, the authors should be careful not to show in tables or figures symmetric error bars that would yield results that are out of range (e.g., negative error rates).
        \item If error bars are reported in tables or plots, the authors should explain in the text how they were calculated and reference the corresponding figures or tables in the text.
    \end{itemize}

\item {\bf Experiments compute resources}
    \item[] Question: For each experiment, does the paper provide sufficient information on the computer resources (type of compute workers, memory, time of execution) needed to reproduce the experiments?
    \item[] Answer: \answerYes{} %
    \item[] Justification: Refer to App.~\ref{app:implementation}.
    \item[] Guidelines:
    \begin{itemize}
        \item The answer \answerNA{} means that the paper does not include experiments.
        \item The paper should indicate the type of compute workers CPU or GPU, internal cluster, or cloud provider, including relevant memory and storage.
        \item The paper should provide the amount of compute required for each of the individual experimental runs as well as estimate the total compute. 
        \item The paper should disclose whether the full research project required more compute than the experiments reported in the paper (e.g., preliminary or failed experiments that didn't make it into the paper). 
    \end{itemize}
    
\item {\bf Code of ethics}
    \item[] Question: Does the research conducted in the paper conform, in every respect, with the NeurIPS Code of Ethics \url{https://neurips.cc/public/EthicsGuidelines}?
    \item[] Answer: \answerYes{} %
    \item[] Justification: The research conducted in the paper conform, in every respect, with the NeurIPS Code of Ethics.
    \item[] Guidelines:
    \begin{itemize}
        \item The answer \answerNA{} means that the authors have not reviewed the NeurIPS Code of Ethics.
        \item If the authors answer \answerNo, they should explain the special circumstances that require a deviation from the Code of Ethics.
        \item The authors should make sure to preserve anonymity (e.g., if there is a special consideration due to laws or regulations in their jurisdiction).
    \end{itemize}

\item {\bf Broader impacts}
    \item[] Question: Does the paper discuss both potential positive societal impacts and negative societal impacts of the work performed?
    \item[] Answer: \answerYes{} %
    \item[] Justification: Refer to Sec.~\ref{sec:conclusion} and our section on ethical considerations in the Appendix.
    \item[] Guidelines:
    \begin{itemize}
        \item The answer \answerNA{} means that there is no societal impact of the work performed.
        \item If the authors answer \answerNA{} or \answerNo, they should explain why their work has no societal impact or why the paper does not address societal impact.
        \item Examples of negative societal impacts include potential malicious or unintended uses (e.g., disinformation, generating fake profiles, surveillance), fairness considerations (e.g., deployment of technologies that could make decisions that unfairly impact specific groups), privacy considerations, and security considerations.
        \item The conference expects that many papers will be foundational research and not tied to particular applications, let alone deployments. However, if there is a direct path to any negative applications, the authors should point it out. For example, it is legitimate to point out that an improvement in the quality of generative models could be used to generate Deepfakes for disinformation. On the other hand, it is not needed to point out that a generic algorithm for optimizing neural networks could enable people to train models that generate Deepfakes faster.
        \item The authors should consider possible harms that could arise when the technology is being used as intended and functioning correctly, harms that could arise when the technology is being used as intended but gives incorrect results, and harms following from (intentional or unintentional) misuse of the technology.
        \item If there are negative societal impacts, the authors could also discuss possible mitigation strategies (e.g., gated release of models, providing defenses in addition to attacks, mechanisms for monitoring misuse, mechanisms to monitor how a system learns from feedback over time, improving the efficiency and accessibility of ML).
    \end{itemize}
    
\item {\bf Safeguards}
    \item[] Question: Does the paper describe safeguards that have been put in place for responsible release of data or models that have a high risk for misuse (e.g., pre-trained language models, image generators, or scraped datasets)?
    \item[] Answer: \answerYes{} %
    \item[] Justification: In \cref{sec:ethical-considerations}, we have included important ethical considerations that users should take note. 
    \item[] Guidelines:
    \begin{itemize}
        \item The answer \answerNA{} means that the paper poses no such risks.
        \item Released models that have a high risk for misuse or dual-use should be released with necessary safeguards to allow for controlled use of the model, for example by requiring that users adhere to usage guidelines or restrictions to access the model or implementing safety filters. 
        \item Datasets that have been scraped from the Internet could pose safety risks. The authors should describe how they avoided releasing unsafe images.
        \item We recognize that providing effective safeguards is challenging, and many papers do not require this, but we encourage authors to take this into account and make a best faith effort.
    \end{itemize}

\item {\bf Licenses for existing assets}
    \item[] Question: Are the creators or original owners of assets (e.g., code, data, models), used in the paper, properly credited and are the license and terms of use explicitly mentioned and properly respected?
    \item[] Answer: \answerYes{} %
    \item[] Justification: The creators or original owners of assets (e.g., code, data, models), used in the paper are properly credited and the license and terms of use are explicitly mentioned and properly respected
    \item[] Guidelines:
    \begin{itemize}
        \item The answer \answerNA{} means that the paper does not use existing assets.
        \item The authors should cite the original paper that produced the code package or dataset.
        \item The authors should state which version of the asset is used and, if possible, include a URL.
        \item The name of the license (e.g., CC-BY 4.0) should be included for each asset.
        \item For scraped data from a particular source (e.g., website), the copyright and terms of service of that source should be provided.
        \item If assets are released, the license, copyright information, and terms of use in the package should be provided. For popular datasets, \url{paperswithcode.com/datasets} has curated licenses for some datasets. Their licensing guide can help determine the license of a dataset.
        \item For existing datasets that are re-packaged, both the original license and the license of the derived asset (if it has changed) should be provided.
        \item If this information is not available online, the authors are encouraged to reach out to the asset's creators.
    \end{itemize}

\item {\bf New assets}
    \item[] Question: Are new assets introduced in the paper well documented and is the documentation provided alongside the assets?
    \item[] Answer: \answerNA{} %
    \item[] Justification: \answerNA{}
    \item[] Guidelines:
    \begin{itemize}
        \item The answer \answerNA{} means that the paper does not release new assets.
        \item Researchers should communicate the details of the dataset\slash code\slash model as part of their submissions via structured templates. This includes details about training, license, limitations, etc. 
        \item The paper should discuss whether and how consent was obtained from people whose asset is used.
        \item At submission time, remember to anonymize your assets (if applicable). You can either create an anonymized URL or include an anonymized zip file.
    \end{itemize}

\item {\bf Crowdsourcing and research with human subjects}
    \item[] Question: For crowdsourcing experiments and research with human subjects, does the paper include the full text of instructions given to participants and screenshots, if applicable, as well as details about compensation (if any)? 
    \item[] Answer: \answerNA{} %
    \item[] Justification: \answerNA{}
    \item[] Guidelines:
    \begin{itemize}
        \item The answer \answerNA{} means that the paper does not involve crowdsourcing nor research with human subjects.
        \item Including this information in the supplemental material is fine, but if the main contribution of the paper involves human subjects, then as much detail as possible should be included in the main paper. 
        \item According to the NeurIPS Code of Ethics, workers involved in data collection, curation, or other labor should be paid at least the minimum wage in the country of the data collector. 
    \end{itemize}

\item {\bf Institutional review board (IRB) approvals or equivalent for research with human subjects}
    \item[] Question: Does the paper describe potential risks incurred by study participants, whether such risks were disclosed to the subjects, and whether Institutional Review Board (IRB) approvals (or an equivalent approval/review based on the requirements of your country or institution) were obtained?
    \item[] Answer: \answerNA{} %
    \item[] Justification: \answerNA{}
    \item[] Guidelines:
    \begin{itemize}
        \item The answer \answerNA{} means that the paper does not involve crowdsourcing nor research with human subjects.
        \item Depending on the country in which research is conducted, IRB approval (or equivalent) may be required for any human subjects research. If you obtained IRB approval, you should clearly state this in the paper. 
        \item We recognize that the procedures for this may vary significantly between institutions and locations, and we expect authors to adhere to the NeurIPS Code of Ethics and the guidelines for their institution. 
        \item For initial submissions, do not include any information that would break anonymity (if applicable), such as the institution conducting the review.
    \end{itemize}

\item {\bf Declaration of LLM usage}
    \item[] Question: Does the paper describe the usage of LLMs if it is an important, original, or non-standard component of the core methods in this research? Note that if the LLM is used only for writing, editing, or formatting purposes and does \emph{not} impact the core methodology, scientific rigor, or originality of the research, declaration is not required.
    \item[] Answer: \answerNA{} %
    \item[] Justification: \answerNA{} 
    \item[] Guidelines:
    \begin{itemize}
        \item The answer \answerNA{} means that the core method development in this research does not involve LLMs as any important, original, or non-standard components.
        \item Please refer to our LLM policy in the NeurIPS handbook for what should or should not be described.
    \end{itemize}

\end{enumerate}

\end{document}